\documentclass[11pt]{article}

\usepackage{acl}

\usepackage{times}
\usepackage{latexsym}
\usepackage{booktabs}
\usepackage{enumitem}

\usepackage[textsize=scriptsize]{todonotes}
\usepackage[T1]{fontenc}

\usepackage[utf8]{inputenc}

\usepackage{microtype}

\usepackage{inconsolata}

\usepackage{graphicx}

\usepackage{amsmath}
\usepackage{cleveref}

\usepackage{amssymb}

\usepackage{adjustbox}
\usepackage{float}

\usepackage{enumitem}

\usepackage{hyperref}

\newcommand{\emoji}[2][]{\includegraphics[height=1em]{#1/#2.png}}

\usepackage{xcolor}         %
\definecolor{ETHPetrol}{RGB}{0,120,148}
\colorlet{MacroColor}{ETHPetrol}

\newcommand{\mm}[1]{{#1}}
\newcommand{\pexp}{\mm{p_{\mathrm{G}}}}
\newcommand{\pama}{\mm{p_{\mathrm{B}}}}
\newcommand{\defeq}{\mathrel{\overset{\raisebox{-0.25ex}{\textnormal{\tiny def}}}{=}}}
\newcommand{\vhead}{\mm{\mathcal{V}_{\text {head}}}}
\newcommand{\CD}{\ensuremath{\mathsf{CD}}}

\newcommand{\CDsymb}{\CD}
\newcommand{\x}{\mm{x_i}}
\newcommand{\xstr}{\mm{\boldsymbol{x}_{<i}}}
\newcommand{\alphabet}{\mm{\Sigma}}
\newcommand{\good}{\ensuremath{\mathsf{GOOD}}\xspace}
\newcommand{\bad}{\ensuremath{\mathsf{BAD}}\xspace}

\newcommand{\mudeltarel}{\ensuremath{\mu_{\Delta \mathrm{REL}}}\xspace}

\usepackage{xspace}

\newcommand{\Baseline}{\textsc{Baseline}\xspace}
\newcommand{\NoContrast}{\textsc{No-Contrast}\xspace}
\newcommand{\NoContrastVHead}{\textsc{No-Contrast–$\vhead$}\xspace}

\usepackage{dblfloatfix}  %
\usepackage[section]{placeins} %

\title{Contrastive Decoding for Synthetic Data Generation\\ in Low-Resource Language Modeling}

\author{Jannek Ulm$^{1}$ \qquad \bf{Kevin Du}$^{1}$ \qquad \textbf{Vésteinn Snæbjarnarson}$^{1,2}$ \\
  $^{1}$ETH Zürich\qquad$^{2}$University of Copenhagen
  \\
  \href{mailto:jannek.ulm@gmail.com}{\texttt{jannek.ulm@gmail.com}} \quad \href{mailto:kevin.du@inf.eth.zh}{\texttt{kevin.du@inf.ethz.ch}} \quad \href{mailto:vest.snae@gmail.com}{\texttt{vest.snae@gmail.com}}
}

\begin{document}
\maketitle

\begin{abstract}
Large language models (LLMs) are trained on huge amounts of textual data, and concerns have been raised that the limits of such data may soon be reached.
A potential solution is to train on synthetic data sampled from LLMs.
In this work, we build on this idea and investigate the benefits of \emph{contrastive decoding} for generating synthetic corpora.
In a controlled setting, we experiment with sampling corpora using the relative difference between a \good and \bad model trained on the same original corpus of 100 million words.
By amplifying the signal from a model that has better performance, we create a synthetic corpus and mix it with the original training data.
Our findings show that training on a mixture of synthesized and real data improves performance on the language modeling objective and a range of downstream tasks.
In particular, we see that training with a mix of synthetic data from contrastive decoding benefits tasks that require more \emph{reasoning skills}, while synthetic data from traditional sampling helps more on tasks dependent on surface-level \emph{linguistic} capabilities.
\vspace{-1.5ex}
\begin{center}
\raisebox{-0.5ex}{\emoji[emoji]{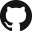}} \url{https://github.com/janulm/CD-for-Synthetic-Data-Generation}
\end{center}
\end{abstract}

% All tables generated by script.

\newcommand{\CommonCaption}{Conventions: values are mean $\pm$ s.e. over $n{=}10$ runs with per-task mean--max checkpointing (see \Cref{sec:eval_statistics}). Arrows indicate directionality; lower Perplexity (TinyBabyLM eval split) is better, higher is better for all other metrics; Reading and Eye Tracking report $\Delta R^{2}$. $^{*}$ marks a significant difference vs.\ \Baseline; percentages in parentheses are relative changes vs.\ \Baseline. \mudeltarel{} is the mean relative improvement across all non-perplexity tasks.}
% Then use:

% --- New Table ---
\newcommand{\tableBaselineExpert}{
\begin{table*}[!t]
\centering
\begin{adjustbox}{max width=\linewidth}
\begin{tabular}{l|llllllll}
\toprule
Name & Perplexity$\downarrow$ & BLiMP$\uparrow$ & BLiMP Supp.$\uparrow$ & Entity Tracking$\uparrow$ & EWoK$\uparrow$ & WUG$\uparrow$ & Reading$\uparrow$ & Eye Tracking$\uparrow$ \\
\midrule
$\good$ & 24.62 & 71.22 & 63.50 & 27.01 & 53.64 & 57.50 & 1.44 & 3.51 \\
\Baseline & 24.46$\pm$0.10 & 71.03$\pm$0.27 & 64.10$\pm$0.60 & 27.82$\pm$1.18 & 53.18$\pm$0.28 & 66.90$\pm$2.47 & 1.76$\pm$0.22 & 3.85$\pm$0.31 \\
\bottomrule
\end{tabular}
\end{adjustbox}

\caption{Reference performance of the \Baseline\ (mean $\pm$ s.e., $n{=}10$ independent runs; per-task mean–max checkpointing per \Cref{sec:eval_statistics}) versus the single fixed $\good$ checkpoint. Because it is a single checkpoint chosen once across seeds rather than per task, it can sit below the \Baseline\ mean on some tasks.}

\label{tab:BaselineExpert}
\end{table*}
}

\newcommand{\tableBaselineExpertShort}{
\begin{table}[]
\centering
\begin{adjustbox}{max width=\linewidth}
\begin{tabular}{l|l}
\toprule
Name & Perplexity$\downarrow$ \\
\midrule
Good & 24.62 \\
\Baseline & 24.46$\pm$0.10 \\
\bottomrule
\end{tabular}
\end{adjustbox}
\caption{\CommonCaption}
\label{tab:BaselineExpertShort}
\end{table}
}

% --- New Table ---
\newcommand{\tableAllModels}{
\begin{table*}[!t] \centering \begin{adjustbox}{max width=0.82\linewidth} \begin{tabular}{l|l|llll} \toprule Name & \mudeltarel$\uparrow$ & Perplexity$\downarrow$ & BLiMP$\uparrow$ & BLiMP Supp.$\uparrow$ & Entity Tracking$\uparrow$ \\ \midrule \Baseline & - & 24.46$\pm$0.10 & 71.03$\pm$0.27 & 64.10$\pm$0.60 & 27.82$\pm$1.18 \\ \midrule No-Contrast-MR-0.3 & 2.96\% & 23.56$\pm$0.11$^{*}$ (3.68\%) & 72.09$\pm$0.17$^{*}$ (1.50\%) & 64.83$\pm$0.73 (1.15\%) & 28.14$\pm$1.75 (1.16\%) \\ No-Contrast-Top-K-100-MR-0.3 & 2.49\% & 23.81$\pm$0.11$^{*}$ (2.66\%) & 72.12$\pm$0.26$^{*}$ (1.55\%) & 64.22$\pm$0.69 (0.19\%) & 28.09$\pm$1.65 (0.98\%) \\ No-Contrast-Top-K-200-MR-0.3 & 3.65\% & 23.65$\pm$0.10$^{*}$ (3.29\%) & 71.78$\pm$0.21$^{*}$ (1.06\%) & 63.98$\pm$0.69 (-0.19\%) & 26.96$\pm$1.23$^{*}$ (-3.08\%) \\ No-Contrast-Top-K-50-MR-0.3 & 3.48\% & 23.88$\pm$0.10$^{*}$ (2.36\%) & 71.52$\pm$0.13$^{*}$ (0.69\%) & 64.45$\pm$0.83 (0.54\%) & 27.23$\pm$1.64$^{*}$ (-2.13\%) \\ No-Contrast-Top-P-90-MR-0.3 & 2.73\% & 23.88$\pm$0.11$^{*}$ (2.37\%) & 71.96$\pm$0.14$^{*}$ (1.31\%) & 64.84$\pm$0.62 (1.16\%) & 26.12$\pm$0.89$^{*}$ (-6.09\%) \\ No-Contrast-Top-P-95-MR-0.3 & 2.33\% & 23.74$\pm$0.12$^{*}$ (2.93\%) & 72.02$\pm$0.22$^{*}$ (1.40\%) & 64.50$\pm$0.63 (0.63\%) & 26.29$\pm$1.31$^{*}$ (-5.51\%) \\ No-Contrast-Top-P-97-MR-0.3 & 2.11\% & 23.61$\pm$0.10$^{*}$ (3.47\%) & 71.62$\pm$0.11$^{*}$ (0.83\%) & 64.33$\pm$0.68 (0.36\%) & 27.29$\pm$1.48$^{*}$ (-1.91\%) \\ No-Contrast-Top-V-Head-MR-0.3 & 0.66\% & 24.33$\pm$0.10$^{*}$ (0.51\%) & 71.67$\pm$0.24$^{*}$ (0.91\%) & 64.86$\pm$0.74 (1.20\%) & 25.47$\pm$1.40$^{*}$ (-8.45\%) \\ \midrule CD-Early-100-MR-0.3 & 2.42\% & 24.02$\pm$0.11$^{*}$ (1.79\%) & 71.31$\pm$0.12$^{*}$ (0.40\%) & 63.54$\pm$0.63 (-0.87\%) & 26.19$\pm$1.51$^{*}$ (-5.87\%) \\ CD-Early-1500-MR-0.3 & 4.26\% & 24.04$\pm$0.14$^{*}$ (1.70\%) & 71.69$\pm$0.26$^{*}$ (0.94\%) & 63.92$\pm$0.57 (-0.28\%) & 27.78$\pm$1.19 (-0.15\%) \\ CD-Early-2000-MR-0.3 & 2.06\% & 24.28$\pm$0.16$^{*}$ (0.73\%) & 71.87$\pm$0.22$^{*}$ (1.19\%) & 63.82$\pm$0.55 (-0.44\%) & 27.55$\pm$1.47 (-0.98\%) \\ \midrule CD-Drop-0.1-MR-0.3 & -1.42\% & 24.02$\pm$0.10$^{*}$ (1.78\%) & 71.55$\pm$0.20$^{*}$ (0.74\%) & 64.39$\pm$0.60 (0.45\%) & 22.59$\pm$0.93$^{*}$ (-18.80\%) \\ CD-Drop-0.3-MR-0.3 & 0.99\% & 24.09$\pm$0.19$^{*}$ (1.52\%) & 71.39$\pm$0.14$^{*}$ (0.52\%) & 64.86$\pm$0.64 (1.19\%) & 24.22$\pm$1.05$^{*}$ (-12.93\%) \\ CD-Drop-0.5-MR-0.3 & 2.52\% & 23.94$\pm$0.10$^{*}$ (2.11\%) & 71.80$\pm$0.28$^{*}$ (1.09\%) & 64.91$\pm$0.60 (1.27\%) & 28.72$\pm$1.00$^{*}$ (3.23\%) \\ CD-Drop-0.7-MR-0.3 & 3.29\% & 24.06$\pm$0.13$^{*}$ (1.65\%) & 71.79$\pm$0.31$^{*}$ (1.08\%) & 65.19$\pm$0.70 (1.71\%) & 28.91$\pm$1.64$^{*}$ (3.91\%) \\ \midrule CD-Small-100-MR-0.3 & 1.65\% & 23.81$\pm$0.14$^{*}$ (2.65\%) & 71.97$\pm$0.27$^{*}$ (1.33\%) & 64.86$\pm$0.58 (1.19\%) & 29.59$\pm$1.14$^{*}$ (6.38\%) \\ CD-Small-10-MR-0.3 & 3.66\% & 23.86$\pm$0.11$^{*}$ (2.44\%) & 71.95$\pm$0.22$^{*}$ (1.30\%) & 64.95$\pm$0.56 (1.33\%) & 27.68$\pm$1.21 (-0.49\%) \\ CD-Small-20-MR-0.3 & 3.55\% & 23.73$\pm$0.14$^{*}$ (2.96\%) & 71.84$\pm$0.19$^{*}$ (1.15\%) & 64.09$\pm$0.66 (-0.01\%) & 29.25$\pm$1.32$^{*}$ (5.15\%) \\ CD-Small-50-MR-0.3 & 2.30\% & 23.73$\pm$0.11$^{*}$ (2.97\%) & 71.97$\pm$0.23$^{*}$ (1.33\%) & \textbf{65.55$\pm$0.58}$^{*}$ (2.27\%) & 29.28$\pm$1.46$^{*}$ (5.26\%) \\ CD-Small-5-MR-0.3 & 2.97\% & 23.97$\pm$0.10$^{*}$ (1.99\%) & 71.46$\pm$0.11$^{*}$ (0.62\%) & 63.89$\pm$0.53 (-0.33\%) & 28.44$\pm$1.03$^{*}$ (2.25\%) \\ \midrule CD-Early-500-MR-0.1 & -0.44\% & 24.11$\pm$0.10$^{*}$ (1.42\%) & 72.11$\pm$0.21$^{*}$ (1.53\%) & 63.59$\pm$0.49 (-0.79\%) & 27.22$\pm$1.16$^{*}$ (-2.17\%) \\ CD-Early-500-MR-0.2 & 3.54\% & 23.64$\pm$0.10$^{*}$ (3.36\%) & \textbf{72.49$\pm$0.18}$^{*}$ (2.06\%) & 64.94$\pm$0.57 (1.31\%) & 31.25$\pm$1.12$^{*}$ (12.34\%) \\ CD-Early-500-MR-0.3 & 4.90\% & 23.73$\pm$0.10$^{*}$ (2.98\%) & 71.72$\pm$0.19$^{*}$ (0.98\%) & 65.10$\pm$0.60$^{*}$ (1.56\%) & 30.38$\pm$0.65$^{*}$ (9.19\%) \\ CD-Early-500-MR-0.4 & 3.54\% & \textbf{23.42$\pm$0.13}$^{*}$ (4.23\%) & 70.90$\pm$0.21 (-0.17\%) & 63.69$\pm$0.55 (-0.63\%) & \textbf{33.30$\pm$0.84}$^{*}$ (19.70\%) \\ CD-Early-500-MR-0.5 & 1.82\% & 23.64$\pm$0.16$^{*}$ (3.35\%) & 69.46$\pm$0.20$^{*}$ (-2.21\%) & 62.84$\pm$0.61$^{*}$ (-1.96\%) & 28.68$\pm$1.31$^{*}$ (3.09\%) \\ CD-Early-500-MR-0.6 & 1.62\% & 23.86$\pm$0.09$^{*}$ (2.43\%) & 68.91$\pm$0.21$^{*}$ (-2.98\%) & 62.30$\pm$0.58$^{*}$ (-2.81\%) & 30.45$\pm$1.09$^{*}$ (9.47\%) \\ CD-Early-500-MR-0.7 & 2.21\% & 25.13$\pm$0.12$^{*}$ (-2.73\%) & 68.18$\pm$0.19$^{*}$ (-4.00\%) & 62.42$\pm$0.67$^{*}$ (-2.62\%) & 31.01$\pm$1.10$^{*}$ (11.48\%) \\ CD-Early-500-MR-0.8 & 0.72\% & 26.14$\pm$0.11$^{*}$ (-6.86\%) & 67.42$\pm$0.25$^{*}$ (-5.07\%) & 61.30$\pm$0.82$^{*}$ (-4.36\%) & 30.57$\pm$0.60$^{*}$ (9.89\%) \\ CD-Early-500-MR-0.9 & 1.16\% & 30.00$\pm$0.12$^{*}$ (-22.64\%) & 66.50$\pm$0.25$^{*}$ (-6.38\%) & 59.86$\pm$0.79$^{*}$ (-6.62\%) & 31.76$\pm$1.11$^{*}$ (14.18\%) \\ \midrule CD-Early-500-Top-K-100-MR-0.3 & 4.90\% & 23.79$\pm$0.12$^{*}$ (2.73\%) & 71.49$\pm$0.18$^{*}$ (0.65\%) & 65.29$\pm$0.80$^{*}$ (1.87\%) & 33.11$\pm$0.62$^{*}$ (19.02\%) \\ CD-Early-500-Top-K-200-MR-0.3 & \textbf{\textbf{5.69\%}} & 23.77$\pm$0.10$^{*}$ (2.80\%) & 71.87$\pm$0.35$^{*}$ (1.19\%) & 64.23$\pm$0.59 (0.20\%) & 31.05$\pm$0.79$^{*}$ (11.61\%) \\ CD-Early-500-Top-K-50-MR-0.3 & 4.64\% & 23.90$\pm$0.12$^{*}$ (2.30\%) & 71.90$\pm$0.21$^{*}$ (1.23\%) & 64.74$\pm$0.68 (1.01\%) & 30.29$\pm$1.49$^{*}$ (8.89\%) \\ CD-Early-500-Top-P-90-MR-0.3 & 4.91\% & 23.74$\pm$0.10$^{*}$ (2.93\%) & 72.16$\pm$0.14$^{*}$ (1.60\%) & 64.69$\pm$0.65 (0.92\%) & 30.43$\pm$1.07$^{*}$ (9.37\%) \\ CD-Early-500-Top-P-95-MR-0.3 & 4.54\% & 23.80$\pm$0.15$^{*}$ (2.69\%) & 71.36$\pm$0.27$^{*}$ (0.47\%) & 64.79$\pm$0.62 (1.09\%) & 32.56$\pm$0.74$^{*}$ (17.06\%) \\ CD-Early-500-Top-P-97-MR-0.3 & 2.98\% & 23.86$\pm$0.13$^{*}$ (2.46\%) & 71.69$\pm$0.20$^{*}$ (0.94\%) & 64.54$\pm$0.56 (0.69\%) & 30.20$\pm$0.92$^{*}$ (8.56\%) \\ \bottomrule \end{tabular} \end{adjustbox}

\begin{adjustbox}{max width=0.82\linewidth} \begin{tabular}{l|l|llll} \toprule Name & \mudeltarel$\uparrow$ & EWoK$\uparrow$ & WUG$\uparrow$ & Reading$\uparrow$ & Eye Tracking$\uparrow$ \\ \midrule \Baseline & - & 53.18$\pm$0.28 & 66.90$\pm$2.47 & 1.76$\pm$0.22 & 3.85$\pm$0.31 \\ \midrule No-Contrast-MR-0.3 & 2.96\% & 53.17$\pm$0.30 (-0.01\%) & 64.67$\pm$1.66$^{*}$ (-3.34\%) & 1.91$\pm$0.25 (8.34\%) & 4.31$\pm$0.33$^{*}$ (11.92\%) \\ No-Contrast-Top-K-100-MR-0.3 & 2.49\% & 53.43$\pm$0.32 (0.48\%) & 66.71$\pm$2.15 (-0.28\%) & 1.85$\pm$0.27 (4.65\%) & 4.23$\pm$0.38 (9.86\%) \\ No-Contrast-Top-K-200-MR-0.3 & 3.65\% & 53.52$\pm$0.32 (0.64\%) & 67.81$\pm$1.53 (1.36\%) & 1.96$\pm$0.26 (10.76\%) & 4.43$\pm$0.35$^{*}$ (15.01\%) \\ No-Contrast-Top-K-50-MR-0.3 & 3.48\% & 53.37$\pm$0.30 (0.37\%) & 67.38$\pm$1.82 (0.71\%) & 1.97$\pm$0.27 (11.83\%) & 4.33$\pm$0.36$^{*}$ (12.38\%) \\ No-Contrast-Top-P-90-MR-0.3 & 2.73\% & 53.36$\pm$0.27 (0.35\%) & 66.25$\pm$2.05 (-0.97\%) & 1.94$\pm$0.23 (9.92\%) & 4.37$\pm$0.32$^{*}$ (13.44\%) \\ No-Contrast-Top-P-95-MR-0.3 & 2.33\% & 53.41$\pm$0.32 (0.45\%) & 66.44$\pm$1.52 (-0.69\%) & 1.90$\pm$0.26 (7.51\%) & 4.33$\pm$0.35$^{*}$ (12.51\%) \\ No-Contrast-Top-P-97-MR-0.3 & 2.11\% & 53.24$\pm$0.28 (0.12\%) & 66.00$\pm$1.63 (-1.35\%) & 1.87$\pm$0.24 (6.20\%) & 4.26$\pm$0.32$^{*}$ (10.51\%) \\ No-Contrast-Top-V-Head-MR-0.3 & 0.66\% & 53.03$\pm$0.31 (-0.27\%) & 66.67$\pm$1.58 (-0.35\%) & 1.76$\pm$0.23 (-0.54\%) & 4.32$\pm$0.33$^{*}$ (12.12\%) \\ \midrule CD-Early-100-MR-0.3 & 2.42\% & 53.19$\pm$0.30 (0.03\%) & 66.83$\pm$1.58 (-0.10\%) & 1.89$\pm$0.25 (7.02\%) & 4.48$\pm$0.34$^{*}$ (16.30\%) \\ CD-Early-1500-MR-0.3 & 4.26\% & 53.61$\pm$0.31 (0.81\%) & 67.89$\pm$2.26 (1.48\%) & \textbf{2.03$\pm$0.26} (14.95\%) & 4.32$\pm$0.34$^{*}$ (12.09\%) \\ CD-Early-2000-MR-0.3 & 2.06\% & 53.30$\pm$0.29 (0.23\%) & 68.67$\pm$1.67 (2.64\%) & 1.80$\pm$0.24 (2.23\%) & 4.22$\pm$0.35 (9.55\%) \\ \midrule CD-Drop-0.1-MR-0.3 & -1.42\% & 53.11$\pm$0.29 (-0.12\%) & 65.33$\pm$2.11 (-2.34\%) & 1.81$\pm$0.24 (2.80\%) & 4.14$\pm$0.32 (7.36\%) \\ CD-Drop-0.3-MR-0.3 & 0.99\% & 53.43$\pm$0.31 (0.48\%) & 67.28$\pm$1.37 (0.56\%) & 1.91$\pm$0.26 (8.15\%) & 4.20$\pm$0.33 (8.95\%) \\ CD-Drop-0.5-MR-0.3 & 2.52\% & 53.17$\pm$0.33 (-0.00\%) & 68.28$\pm$1.74 (2.06\%) & 1.75$\pm$0.24 (-0.72\%) & 4.27$\pm$0.33 (10.74\%) \\ CD-Drop-0.7-MR-0.3 & 3.29\% & 53.62$\pm$0.40 (0.83\%) & 66.80$\pm$1.72 (-0.15\%) & 1.90$\pm$0.35 (7.76\%) & 4.16$\pm$0.44 (7.92\%) \\ \midrule CD-Small-100-MR-0.3 & 1.65\% & 53.36$\pm$0.28 (0.34\%) & 66.20$\pm$1.61 (-1.05\%) & 1.68$\pm$0.21 (-4.65\%) & 4.16$\pm$0.31 (7.99\%) \\ CD-Small-10-MR-0.3 & 3.66\% & 53.50$\pm$0.32 (0.61\%) & 68.80$\pm$2.24 (2.84\%) & 1.93$\pm$0.24 (9.12\%) & 4.27$\pm$0.31$^{*}$ (10.87\%) \\ CD-Small-20-MR-0.3 & 3.55\% & 53.45$\pm$0.27 (0.50\%) & 69.05$\pm$2.53 (3.21\%) & 1.79$\pm$0.22 (1.59\%) & 4.37$\pm$0.31$^{*}$ (13.29\%) \\ CD-Small-50-MR-0.3 & 2.30\% & 53.29$\pm$0.28 (0.20\%) & 66.45$\pm$1.54 (-0.67\%) & 1.78$\pm$0.23 (1.13\%) & 4.10$\pm$0.31 (6.54\%) \\ CD-Small-5-MR-0.3 & 2.97\% & 53.23$\pm$0.30 (0.09\%) & 67.40$\pm$1.37 (0.75\%) & 1.83$\pm$0.22 (3.74\%) & 4.38$\pm$0.32$^{*}$ (13.68\%) \\ \midrule CD-Early-500-MR-0.1 & -0.44\% & 53.45$\pm$0.33 (0.51\%) & 66.10$\pm$1.44 (-1.20\%) & 1.69$\pm$0.22 (-4.08\%) & 3.97$\pm$0.32 (3.09\%) \\ CD-Early-500-MR-0.2 & 3.54\% & 53.53$\pm$0.27 (0.66\%) & 66.35$\pm$1.48 (-0.82\%) & 1.78$\pm$0.23 (0.79\%) & 4.18$\pm$0.31 (8.41\%) \\ CD-Early-500-MR-0.3 & 4.90\% & 53.80$\pm$0.29$^{*}$ (1.18\%) & \textbf{70.55$\pm$2.32}$^{*}$ (5.46\%) & 1.79$\pm$0.22 (1.30\%) & 4.42$\pm$0.32$^{*}$ (14.64\%) \\ CD-Early-500-MR-0.4 & 3.54\% & 53.41$\pm$0.28 (0.44\%) & 66.50$\pm$1.87 (-0.60\%) & 1.74$\pm$0.23 (-1.19\%) & 4.13$\pm$0.31 (7.27\%) \\ CD-Early-500-MR-0.5 & 1.82\% & 53.36$\pm$0.29 (0.34\%) & 67.00$\pm$1.76 (0.15\%) & 1.77$\pm$0.22 (0.34\%) & 4.35$\pm$0.32 (12.98\%) \\ CD-Early-500-MR-0.6 & 1.62\% & 53.04$\pm$0.31 (-0.25\%) & 68.35$\pm$1.89 (2.17\%) & 1.69$\pm$0.22 (-4.36\%) & 4.24$\pm$0.32 (10.10\%) \\ CD-Early-500-MR-0.7 & 2.21\% & 52.91$\pm$0.28 (-0.50\%) & 64.75$\pm$1.85 (-3.21\%) & 1.82$\pm$0.23 (2.95\%) & 4.29$\pm$0.34 (11.37\%) \\ CD-Early-500-MR-0.8 & 0.72\% & 52.73$\pm$0.31 (-0.85\%) & 64.28$\pm$1.89$^{*}$ (-3.92\%) & 1.80$\pm$0.24 (1.79\%) & 4.15$\pm$0.32 (7.59\%) \\ CD-Early-500-MR-0.9 & 1.16\% & 52.57$\pm$0.31 (-1.13\%) & 65.06$\pm$1.23 (-2.76\%) & 1.79$\pm$0.25 (1.35\%) & 4.22$\pm$0.34 (9.50\%) \\ \midrule CD-Early-500-Top-K-100-MR-0.3 & 4.90\% & \textbf{53.94$\pm$0.36}$^{*}$ (1.44\%) & 67.44$\pm$2.13 (0.80\%) & 1.73$\pm$0.26 (-2.20\%) & 4.34$\pm$0.35$^{*}$ (12.74\%) \\ CD-Early-500-Top-K-200-MR-0.3 & \textbf{\textbf{5.69\%}} & 53.61$\pm$0.31 (0.82\%) & 67.90$\pm$2.00 (1.49\%) & 1.92$\pm$0.25 (8.73\%) & 4.46$\pm$0.33$^{*}$ (15.78\%) \\ CD-Early-500-Top-K-50-MR-0.3 & 4.64\% & 53.47$\pm$0.32 (0.55\%) & 67.56$\pm$2.48 (0.99\%) & 1.93$\pm$0.26 (9.49\%) & 4.25$\pm$0.35 (10.30\%) \\ CD-Early-500-Top-P-90-MR-0.3 & 4.91\% & 53.68$\pm$0.30 (0.94\%) & 68.35$\pm$1.44 (2.17\%) & 1.85$\pm$0.23 (4.59\%) & 4.42$\pm$0.33$^{*}$ (14.77\%) \\ CD-Early-500-Top-P-95-MR-0.3 & 4.54\% & 53.60$\pm$0.29 (0.80\%) & 65.78$\pm$1.99 (-1.68\%) & 1.82$\pm$0.24 (2.86\%) & 4.28$\pm$0.33 (11.14\%) \\ CD-Early-500-Top-P-97-MR-0.3 & 2.98\% & 53.63$\pm$0.30 (0.86\%) & 64.85$\pm$1.50 (-3.06\%) & 1.82$\pm$0.22 (3.06\%) & 4.23$\pm$0.31 (9.84\%) \\ \bottomrule
\end{tabular}
\end{adjustbox} 

\caption{Full sweep of all experiments. Naming scheme: \NoContrast{} = ancestral sampling from $\pexp$; \NoContrastVHead{} = ancestral sampling restricted to the $\alpha$-head $\vhead(\cdot)$ of $\pexp$; \textsc{CD--Early--\emph{k}} = contrastive decoding with the amateur $\pama$ taken as an earlier training checkpoint at step \emph{k}; \textsc{CD--Small--\emph{r}} = $\pama$ is a smaller model (about $\emph{r}\times$ fewer parameters than $\pexp$); \textsc{CD--Drop--\emph{p}} = $\pama$ is $\pexp$ run with attention dropout rate \emph{p} at inference; \textsc{CD--Synth--Ratio--\emph{q}} = training mixture uses synthetic fraction \emph{q}. \texttt{G2500} denotes the fixed \good{} checkpoint used for generation (selected at training step 2500). Other conventions follow the universal caption: means $\pm$ s.e.; $^{*}$ indicates significance vs.\ \Baseline; parentheses give relative change vs.\ \Baseline; \mudeltarel{} averages non-perplexity tasks; Reading/Eye Tracking values are the \% increase in variance explained after adding the LM features.}
\label{tab:AllModels}
\end{table*}
}

\newcommand{\tableAllModelsShort}{
\begin{table}[]
\centering
\begin{adjustbox}{max width=\linewidth}
\begin{tabular}{l|l|l}
\toprule
Name & \mudeltarel$\uparrow$ & Perplexity$\downarrow$ \\
\midrule
\Baseline & - & 24.46$\pm$0.10 \\
g2500\_1000\_cs1\_mr03 & 2.42\% & 24.02$\pm$0.11$^{*}$ (1.79\%) \\
g2500\_1500\_cs1\_mr03 & 4.26\% & 24.04$\pm$0.14$^{*}$ (1.70\%) \\
g2500\_2000\_cs1\_mr03 & 2.06\% & 24.28$\pm$0.16$^{*}$ (0.73\%) \\
g2500\_500\_cs1\_mr01 & -0.44\% & 24.11$\pm$0.10$^{*}$ (1.42\%) \\
g2500\_500\_cs1\_mr02 & 3.54\% & 23.64$\pm$0.10$^{*}$ (3.36\%) \\
g2500\_500\_cs1\_mr03 & \textbf{\textbf{4.90\%}} & 23.73$\pm$0.10$^{*}$ (2.98\%) \\
g2500\_500\_cs1\_mr04 & 3.54\% & \textbf{23.42$\pm$0.13}$^{*}$ (4.23\%) \\
g2500\_500\_cs1\_mr05 & 1.82\% & 23.64$\pm$0.16$^{*}$ (3.35\%) \\
g2500\_500\_tail\_cs1\_mr03 & 2.07\% & 23.50$\pm$0.13$^{*}$ (3.92\%) \\
g2500\_drop\_01\_cs1\_mr03 & -1.42\% & 24.02$\pm$0.10$^{*}$ (1.78\%) \\
g2500\_drop\_03\_cs1\_mr03 & 0.99\% & 24.09$\pm$0.19$^{*}$ (1.52\%) \\
g2500\_drop\_05\_cs1\_mr03 & 2.52\% & 23.94$\pm$0.10$^{*}$ (2.11\%) \\
g2500\_drop\_07\_cs1\_mr03 & 3.29\% & 24.06$\pm$0.13$^{*}$ (1.65\%) \\
g2500\_no\_contrast\_mr03 & 2.96\% & 23.56$\pm$0.11$^{*}$ (3.68\%) \\
g2500\_no\_contrast\_v\_head\_mr03 & 0.66\% & 24.33$\pm$0.10$^{*}$ (0.51\%) \\
g2500\_small\_100\_cs1\_mr03 & 1.65\% & 23.81$\pm$0.14$^{*}$ (2.65\%) \\
g2500\_small\_10\_cs1\_mr03 & 3.66\% & 23.86$\pm$0.11$^{*}$ (2.44\%) \\
g2500\_small\_20\_cs1\_mr03 & 3.55\% & 23.73$\pm$0.14$^{*}$ (2.96\%) \\
g2500\_small\_50\_cs1\_mr03 & 2.30\% & 23.73$\pm$0.11$^{*}$ (2.97\%) \\
g2500\_small\_5\_cs1\_mr03 & 2.97\% & 23.97$\pm$0.10$^{*}$ (1.99\%) \\
\bottomrule
\end{tabular}
\end{adjustbox}
\caption{All Models. Statistical significance is marked with $^{*}$ and the percentages are the relative change.}
\label{tab:AllModelsShort}
\end{table}
}

% --- New Table ---
\newcommand{\tableModelSize}{
\begin{table*}[!t]
\centering
\begin{adjustbox}{max width=\linewidth}
\begin{tabular}{l|l|llllllll}
\toprule
Name & \mudeltarel$\uparrow$ & Perplexity$\downarrow$ & BLiMP$\uparrow$ & BLiMP Supp.$\uparrow$ & Entity Tracking$\uparrow$ & EWoK$\uparrow$ & WUG$\uparrow$ & Reading$\uparrow$ & Eye Tracking$\uparrow$ \\
\midrule
\Baseline & - & 24.46$\pm$0.10 & 71.03$\pm$0.27 & 64.10$\pm$0.60 & 27.82$\pm$1.18 & 53.18$\pm$0.28 & 66.90$\pm$2.47 & 1.76$\pm$0.22 & 3.85$\pm$0.31 \\
CD-Small-10 & \textbf{\textbf{3.66\%}} & 23.86$\pm$0.11$^{*}$ (2.44\%) & 71.95$\pm$0.22$^{*}$ (1.30\%) & 64.95$\pm$0.56 (1.33\%) & 27.68$\pm$1.21 (-0.49\%) & \textbf{53.50$\pm$0.32} (0.61\%) & 68.80$\pm$2.24 (2.84\%) & \textbf{1.93$\pm$0.24} (9.12\%) & 4.27$\pm$0.31$^{*}$ (10.87\%) \\
CD-Small-20 & 3.55\% & 23.73$\pm$0.14$^{*}$ (2.96\%) & 71.84$\pm$0.19$^{*}$ (1.15\%) & 64.09$\pm$0.66 (-0.01\%) & 29.25$\pm$1.32$^{*}$ (5.15\%) & 53.45$\pm$0.27 (0.50\%) & \textbf{69.05$\pm$2.53} (3.21\%) & 1.79$\pm$0.22 (1.59\%) & \textbf{4.37$\pm$0.31}$^{*}$ (13.29\%) \\
CD-Small-50 & 2.30\% & \textbf{23.73$\pm$0.11}$^{*}$ (2.97\%) & \textbf{71.97$\pm$0.23}$^{*}$ (1.33\%) & \textbf{65.55$\pm$0.58}$^{*}$ (2.27\%) & 29.28$\pm$1.46$^{*}$ (5.26\%) & 53.29$\pm$0.28 (0.20\%) & 66.45$\pm$1.54 (-0.67\%) & 1.78$\pm$0.23 (1.13\%) & 4.10$\pm$0.31 (6.54\%) \\
CD-Small-100 & 1.65\% & 23.81$\pm$0.14$^{*}$ (2.65\%) & 71.97$\pm$0.27$^{*}$ (1.33\%) & 64.86$\pm$0.58 (1.19\%) & \textbf{29.59$\pm$1.14}$^{*}$ (6.38\%) & 53.36$\pm$0.28 (0.34\%) & 66.20$\pm$1.61 (-1.05\%) & 1.68$\pm$0.21 (-4.65\%) & 4.16$\pm$0.31 (7.99\%) \\
\bottomrule
\end{tabular}
\end{adjustbox}
\caption{\CommonCaption}
\label{tab:ModelSize}
\end{table*}
}

\newcommand{\tableModelSizeShort}{
\begin{table}[]
\centering
\begin{adjustbox}{max width=\linewidth}
\begin{tabular}{l|l|l}
\toprule
Name & \mudeltarel$\uparrow$ & Perplexity$\downarrow$ \\
\midrule
\Baseline & - & 24.46$\pm$0.10 \\
CD-Small-10 & \textbf{\textbf{3.66\%}} & 23.86$\pm$0.11$^{*}$ (2.44\%) \\
CD-Small-20 & 3.55\% & 23.73$\pm$0.14$^{*}$ (2.96\%) \\
CD-Small-50 & 2.30\% & \textbf{23.73$\pm$0.11}$^{*}$ (2.97\%) \\
CD-Small-100 & 1.65\% & 23.81$\pm$0.14$^{*}$ (2.65\%) \\
\bottomrule
\end{tabular}
\end{adjustbox}
\caption{\CommonCaption}
\label{tab:ModelSizeShort}
\end{table}
}

% --- New Table ---
\newcommand{\tableNoContrast}{
\begin{table*}[!t]
\centering
\begin{adjustbox}{max width=\linewidth}
\begin{tabular}{l|l|llllllll}
\toprule
Name & \mudeltarel$\uparrow$ & Perplexity$\downarrow$ & BLiMP$\uparrow$ & BLiMP Supp.$\uparrow$ & Entity Tracking$\uparrow$ & EWoK$\uparrow$ & WUG$\uparrow$ & Reading$\uparrow$ & Eye Tracking$\uparrow$ \\
\midrule
\Baseline & - & 24.46$\pm$0.10 & 71.03$\pm$0.27 & 64.10$\pm$0.60 & 27.82$\pm$1.18 & \textbf{53.18$\pm$0.28} & \textbf{66.90$\pm$2.47} & 1.76$\pm$0.22 & 3.85$\pm$0.31 \\
\NoContrast & \textbf{\textbf{2.96\%}} & \textbf{23.56$\pm$0.11}$^{*}$ (3.68\%) & \textbf{72.09$\pm$0.17}$^{*}$ (1.50\%) & 64.83$\pm$0.73 (1.15\%) & \textbf{28.14$\pm$1.75} (1.16\%) & 53.17$\pm$0.30 (-0.01\%) & 64.67$\pm$1.66$^{*}$ (-3.34\%) & \textbf{1.91$\pm$0.25} (8.34\%) & 4.31$\pm$0.33$^{*}$ (11.92\%) \\
\NoContrastVHead & 0.66\% & 24.33$\pm$0.10$^{*}$ (0.51\%) & 71.67$\pm$0.24$^{*}$ (0.91\%) & \textbf{64.86$\pm$0.74} (1.20\%) & 25.47$\pm$1.40$^{*}$ (-8.45\%) & 53.03$\pm$0.31 (-0.27\%) & 66.67$\pm$1.58 (-0.35\%) & 1.76$\pm$0.23 (-0.54\%) & \textbf{4.32$\pm$0.33}$^{*}$ (12.12\%) \\
\bottomrule
\end{tabular}
\end{adjustbox}
\caption{\CommonCaption}
\label{tab:NoContrast}
\end{table*}
}

\newcommand{\tableNoContrastShort}{
\begin{table}[]
\centering
\begin{adjustbox}{max width=\linewidth}
\begin{tabular}{l|l|l}
\toprule
Name & \mudeltarel$\uparrow$ & Perplexity$\downarrow$ \\
\midrule
Baseline & - & 24.46$\pm$0.10 \\
\NoContrast & \textbf{\textbf{2.96\%}} & \textbf{23.56$\pm$0.11}$^{*}$ (3.68\%) \\
\NoContrastVHead & 0.66\% & 24.33$\pm$0.10$^{*}$ (0.51\%) \\
\bottomrule
\end{tabular}
\end{adjustbox}
\caption{\CommonCaption}
\label{tab:NoContrastShort}
\end{table}
}

% --- New Table ---
\newcommand{\tableDropout}{
\begin{table*}[!t]
\centering
\begin{adjustbox}{max width=\linewidth}
\begin{tabular}{l|l|llllllll}
\toprule
Name & \mudeltarel$\uparrow$ & Perplexity$\downarrow$ & BLiMP$\uparrow$ & BLiMP Supp.$\uparrow$ & Entity Tracking$\uparrow$ & EWoK$\uparrow$ & WUG$\uparrow$ & Reading$\uparrow$ & Eye Tracking$\uparrow$ \\
\midrule
\Baseline & - & 24.46$\pm$0.10 & 71.03$\pm$0.27 & 64.10$\pm$0.60 & 27.82$\pm$1.18 & 53.18$\pm$0.28 & 66.90$\pm$2.47 & 1.76$\pm$0.22 & 3.85$\pm$0.31 \\
CD-Drop-0.1 & -1.42\% & 24.02$\pm$0.10$^{*}$ (1.78\%) & 71.55$\pm$0.20$^{*}$ (0.74\%) & 64.39$\pm$0.60 (0.45\%) & 22.59$\pm$0.93$^{*}$ (-18.80\%) & 53.11$\pm$0.29 (-0.12\%) & 65.33$\pm$2.11 (-2.34\%) & 1.81$\pm$0.24 (2.80\%) & 4.14$\pm$0.32 (7.36\%) \\
CD-Drop-0.3 & 0.99\% & 24.09$\pm$0.19$^{*}$ (1.52\%) & 71.39$\pm$0.14$^{*}$ (0.52\%) & 64.86$\pm$0.64 (1.19\%) & 24.22$\pm$1.05$^{*}$ (-12.93\%) & 53.43$\pm$0.31 (0.48\%) & 67.28$\pm$1.37 (0.56\%) & \textbf{1.91$\pm$0.26} (8.15\%) & 4.20$\pm$0.33 (8.95\%) \\
CD-Drop-0.5 & 2.52\% & \textbf{23.94$\pm$0.10}$^{*}$ (2.11\%) & \textbf{71.80$\pm$0.28}$^{*}$ (1.09\%) & 64.91$\pm$0.60 (1.27\%) & 28.72$\pm$1.00$^{*}$ (3.23\%) & 53.17$\pm$0.33 (-0.00\%) & \textbf{68.28$\pm$1.74} (2.06\%) & 1.75$\pm$0.24 (-0.72\%) & \textbf{4.27$\pm$0.33} (10.74\%) \\
CD-Drop-0.7 & \textbf{\textbf{3.29\%}} & 24.06$\pm$0.13$^{*}$ (1.65\%) & 71.79$\pm$0.31$^{*}$ (1.08\%) & \textbf{65.19$\pm$0.70} (1.71\%) & \textbf{28.91$\pm$1.64}$^{*}$ (3.91\%) & \textbf{53.62$\pm$0.40} (0.83\%) & 66.80$\pm$1.72 (-0.15\%) & 1.90$\pm$0.35 (7.76\%) & 4.16$\pm$0.44 (7.92\%) \\
\bottomrule
\end{tabular}
\end{adjustbox}
\caption{\CommonCaption}
\label{tab:Dropout}
\end{table*}
}

\newcommand{\tableDropoutShort}{
\begin{table}[]
\centering
\begin{adjustbox}{max width=\linewidth}
\begin{tabular}{l|l|l}
\toprule
Name & \mudeltarel$\uparrow$ & Perplexity$\downarrow$ \\
\midrule
\Baseline & - & 24.46$\pm$0.10 \\
CD-Drop-0.1 & -1.42\% & 24.02$\pm$0.10$^{*}$ (1.78\%) \\
CD-Drop-0.3 & 0.99\% & 24.09$\pm$0.19$^{*}$ (1.52\%) \\
CD-Drop-0.5 & 2.52\% & \textbf{23.94$\pm$0.10}$^{*}$ (2.11\%) \\
CD-Drop-0.7 & \textbf{\textbf{3.29\%}} & 24.06$\pm$0.13$^{*}$ (1.65\%) \\
\bottomrule
\end{tabular}
\end{adjustbox}
\caption{\CommonCaption}
\label{tab:DropoutShort}
\end{table}
}

% --- New Table ---
\newcommand{\tableEarlySteps}{
\begin{table*}[!t]
\centering
\begin{adjustbox}{max width=\linewidth}
\begin{tabular}{l|l|llllllll}
\toprule
Name & \mudeltarel$\uparrow$ & Perplexity$\downarrow$ & BLiMP$\uparrow$ & BLiMP Supp.$\uparrow$ & Entity Tracking$\uparrow$ & EWoK$\uparrow$ & WUG$\uparrow$ & Reading$\uparrow$ & Eye Tracking$\uparrow$ \\
\midrule
\Baseline & - & 24.46$\pm$0.10 & 71.03$\pm$0.27 & 64.10$\pm$0.60 & 27.82$\pm$1.18 & 53.18$\pm$0.28 & 66.90$\pm$2.47 & 1.76$\pm$0.22 & 3.85$\pm$0.31 \\
CD-Early-500 & \textbf{\textbf{4.90\%}} & \textbf{23.73$\pm$0.10}$^{*}$ (2.98\%) & 71.72$\pm$0.19$^{*}$ (0.98\%) & \textbf{65.10$\pm$0.60}$^{*}$ (1.56\%) & \textbf{30.38$\pm$0.65}$^{*}$ (9.19\%) & \textbf{53.80$\pm$0.29}$^{*}$ (1.18\%) & \textbf{70.55$\pm$2.32}$^{*}$ (5.46\%) & 1.79$\pm$0.22 (1.30\%) & 4.42$\pm$0.32$^{*}$ (14.64\%) \\
CD-Early-1000 & 2.42\% & 24.02$\pm$0.11$^{*}$ (1.79\%) & 71.31$\pm$0.12$^{*}$ (0.40\%) & 63.54$\pm$0.63 (-0.87\%) & 26.19$\pm$1.51$^{*}$ (-5.87\%) & 53.19$\pm$0.30 (0.03\%) & 66.83$\pm$1.58 (-0.10\%) & 1.89$\pm$0.25 (7.02\%) & 4.48$\pm$0.34$^{*}$ (16.30\%) \\
CD-Early-1500 & 4.26\% & 24.04$\pm$0.14$^{*}$ (1.70\%) & 71.69$\pm$0.26$^{*}$ (0.94\%) & 63.92$\pm$0.57 (-0.28\%) & 27.78$\pm$1.19 (-0.15\%) & 53.61$\pm$0.31 (0.81\%) & 67.89$\pm$2.26 (1.48\%) & \textbf{2.03$\pm$0.26} (14.95\%) & 4.32$\pm$0.34$^{*}$ (12.09\%) \\
CD-Early-2000 & 2.06\% & 24.28$\pm$0.16$^{*}$ (0.73\%) & \textbf{71.87$\pm$0.22}$^{*}$ (1.19\%) & 63.82$\pm$0.55 (-0.44\%) & 27.55$\pm$1.47 (-0.98\%) & 53.30$\pm$0.29 (0.23\%) & 68.67$\pm$1.67 (2.64\%) & 1.80$\pm$0.24 (2.23\%) & 4.22$\pm$0.35 (9.55\%) \\
\bottomrule
\end{tabular}
\end{adjustbox}
\caption{\CommonCaption}
\label{tab:EarlySteps}
\end{table*}
}

\newcommand{\tableEarlyStepsShort}{
\begin{table}[]
\centering
\begin{adjustbox}{max width=\linewidth}
\begin{tabular}{l|l|l}
\toprule
Name & \mudeltarel$\uparrow$ & Perplexity$\downarrow$ \\
\midrule
\Baseline & - & 24.46$\pm$0.10 \\
CD-Early-500 & \textbf{\textbf{4.90\%}} & \textbf{23.73$\pm$0.10}$^{*}$ (2.98\%) \\
CD-Early-1000 & 2.42\% & 24.02$\pm$0.11$^{*}$ (1.79\%) \\
CD-Early-1500 & 4.26\% & 24.04$\pm$0.14$^{*}$ (1.70\%) \\
CD-Early-2000 & 2.06\% & 24.28$\pm$0.16$^{*}$ (0.73\%) \\
\bottomrule
\end{tabular}
\end{adjustbox}
\caption{\CommonCaption}
\label{tab:EarlyStepsShort}
\end{table}
}

% --- New Table ---
\newcommand{\tableMixingRatio}{
\begin{table*}[!t]
\centering
\begin{adjustbox}{max width=\linewidth}
\begin{tabular}{l|l|llllllll}
\toprule
Name & \mudeltarel$\uparrow$ & Perplexity$\downarrow$ & BLiMP$\uparrow$ & BLiMP Supp.$\uparrow$ & Entity Tracking$\uparrow$ & EWoK$\uparrow$ & WUG$\uparrow$ & Reading$\uparrow$ & Eye Tracking$\uparrow$ \\
\midrule
\Baseline & - & 24.46$\pm$0.10 & 71.03$\pm$0.27 & 64.10$\pm$0.60 & 27.82$\pm$1.18 & 53.18$\pm$0.28 & 66.90$\pm$2.47 & 1.76$\pm$0.22 & 3.85$\pm$0.31 \\
CD-Synth-Ratio-0.1 & -0.44\% & 24.11$\pm$0.10$^{*}$ (1.42\%) & 72.11$\pm$0.21$^{*}$ (1.53\%) & 63.59$\pm$0.49 (-0.79\%) & 27.22$\pm$1.16$^{*}$ (-2.17\%) & 53.45$\pm$0.33 (0.51\%) & 66.10$\pm$1.44 (-1.20\%) & 1.69$\pm$0.22 (-4.08\%) & 3.97$\pm$0.32 (3.09\%) \\
CD-Synth-Ratio-0.2 & 3.54\% & 23.64$\pm$0.10$^{*}$ (3.36\%) & \textbf{72.49$\pm$0.18}$^{*}$ (2.06\%) & 64.94$\pm$0.57 (1.31\%) & 31.25$\pm$1.12$^{*}$ (12.34\%) & 53.53$\pm$0.27 (0.66\%) & 66.35$\pm$1.48 (-0.82\%) & 1.78$\pm$0.23 (0.79\%) & 4.18$\pm$0.31 (8.41\%) \\
CD-Synth-Ratio-0.3 & \textbf{\textbf{4.90\%}} & 23.73$\pm$0.10$^{*}$ (2.98\%) & 71.72$\pm$0.19$^{*}$ (0.98\%) & \textbf{65.10$\pm$0.60}$^{*}$ (1.56\%) & 30.38$\pm$0.65$^{*}$ (9.19\%) & \textbf{53.80$\pm$0.29}$^{*}$ (1.18\%) & \textbf{70.55$\pm$2.32}$^{*}$ (5.46\%) & 1.79$\pm$0.22 (1.30\%) & \textbf{4.42$\pm$0.32}$^{*}$ (14.64\%) \\
CD-Synth-Ratio-0.4 & 3.54\% & \textbf{23.42$\pm$0.13}$^{*}$ (4.23\%) & 70.90$\pm$0.21 (-0.17\%) & 63.69$\pm$0.55 (-0.63\%) & \textbf{33.30$\pm$0.84}$^{*}$ (19.70\%) & 53.41$\pm$0.28 (0.44\%) & 66.50$\pm$1.87 (-0.60\%) & 1.74$\pm$0.23 (-1.19\%) & 4.13$\pm$0.31 (7.27\%) \\
CD-Synth-Ratio-0.5 & 1.82\% & 23.64$\pm$0.16$^{*}$ (3.35\%) & 69.46$\pm$0.20$^{*}$ (-2.21\%) & 62.84$\pm$0.61$^{*}$ (-1.96\%) & 28.68$\pm$1.31$^{*}$ (3.09\%) & 53.36$\pm$0.29 (0.34\%) & 67.00$\pm$1.76 (0.15\%) & 1.77$\pm$0.22 (0.34\%) & 4.35$\pm$0.32 (12.98\%) \\
CD-Synth-Ratio-0.6 & 1.62\% & 23.86$\pm$0.09$^{*}$ (2.43\%) & 68.91$\pm$0.21$^{*}$ (-2.98\%) & 62.30$\pm$0.58$^{*}$ (-2.81\%) & 30.45$\pm$1.09$^{*}$ (9.47\%) & 53.04$\pm$0.31 (-0.25\%) & 68.35$\pm$1.89 (2.17\%) & 1.69$\pm$0.22 (-4.36\%) & 4.24$\pm$0.32 (10.10\%) \\
CD-Synth-Ratio-0.7 & 2.21\% & 25.13$\pm$0.12$^{*}$ (-2.73\%) & 68.18$\pm$0.19$^{*}$ (-4.00\%) & 62.42$\pm$0.67$^{*}$ (-2.62\%) & 31.01$\pm$1.10$^{*}$ (11.48\%) & 52.91$\pm$0.28 (-0.50\%) & 64.75$\pm$1.85 (-3.21\%) & \textbf{1.82$\pm$0.23} (2.95\%) & 4.29$\pm$0.34 (11.37\%) \\
CD-Synth-Ratio-0.8 & 0.72\% & 26.14$\pm$0.11$^{*}$ (-6.86\%) & 67.42$\pm$0.25$^{*}$ (-5.07\%) & 61.30$\pm$0.82$^{*}$ (-4.36\%) & 30.57$\pm$0.60$^{*}$ (9.89\%) & 52.73$\pm$0.31 (-0.85\%) & 64.28$\pm$1.89$^{*}$ (-3.92\%) & 1.80$\pm$0.24 (1.79\%) & 4.15$\pm$0.32 (7.59\%) \\
CD-Synth-Ratio-0.9 & 1.16\% & 30.00$\pm$0.12$^{*}$ (-22.64\%) & 66.50$\pm$0.25$^{*}$ (-6.38\%) & 59.86$\pm$0.79$^{*}$ (-6.62\%) & 31.76$\pm$1.11$^{*}$ (14.18\%) & 52.57$\pm$0.31 (-1.13\%) & 65.06$\pm$1.23 (-2.76\%) & 1.79$\pm$0.25 (1.35\%) & 4.22$\pm$0.34 (9.50\%) \\

\bottomrule
\end{tabular}
\end{adjustbox}
\caption{\CommonCaption}
\label{tab:MixingRatio}
\end{table*}
}

\newcommand{\tableMixingRatioShort}{
\begin{table}[]
\centering
\begin{adjustbox}{max width=\linewidth}
\begin{tabular}{l|l|l}
\toprule
Name & \mudeltarel$\uparrow$ & Perplexity$\downarrow$ \\
\midrule
\Baseline & - & 24.46$\pm$0.10 \\
CD-Synth-Ratio-0.1 & -0.44\% & 24.11$\pm$0.10$^{*}$ (1.42\%) \\
CD-Synth-Ratio-0.2 & 3.54\% & 23.64$\pm$0.10$^{*}$ (3.36\%) \\
CD-Synth-Ratio-0.3 & \textbf{\textbf{4.90\%}} & 23.73$\pm$0.10$^{*}$ (2.98\%) \\
CD-Synth-Ratio-0.4 & 3.54\% & \textbf{23.42$\pm$0.13}$^{*}$ (4.23\%) \\
CD-Synth-Ratio-0.5 & 1.82\% & 23.64$\pm$0.16$^{*}$ (3.35\%) \\
CD-Synth-Ratio-0.6 & 1.62\% & 23.86$\pm$0.09$^{*}$ (2.43\%) \\
CD-Synth-Ratio-0.7 & 2.21\% & 25.13$\pm$0.12$^{*}$ (-2.73\%) \\
CD-Synth-Ratio-0.8 & 0.72\% & 26.14$\pm$0.11$^{*}$ (-6.86\%) \\
CD-Synth-Ratio-0.9 & 1.16\% & 30.00$\pm$0.12$^{*}$ (-22.64\%) \\
\bottomrule
\end{tabular}
\end{adjustbox}

\caption{Mixing ratio ablation for CD-generated synthetic data (CD-Early-500). The ratio indicates the fraction of synthetic data in training batches. \mudeltarel{} is the mean relative improvement over \Baseline{} across non-perplexity tasks; Perplexity shows mean $\pm$ s.e., with parentheses giving relative change vs.\ \Baseline{}; $^{*}$ marks significance vs.\ \Baseline{}. A 30\% mix yields the best overall \mudeltarel{} (+4.90\%), while 40\% attains the lowest perplexity (23.42).}

\label{tab:MixingRatioShort}
\end{table}
}

% --- New Table ---
\newcommand{\tableSummaryExperiments}{
\begin{table*}[!tb]
\centering
\begin{adjustbox}{max width=\linewidth}
\begin{tabular}{l|l|llllllll}
\toprule
Name & \mudeltarel$\uparrow$ & Perplexity$\downarrow$ & BLiMP$\uparrow$ & BLiMP Supp.$\uparrow$ & Entity Tracking$\uparrow$ & EWoK$\uparrow$ & WUG$\uparrow$ & Reading$\uparrow$ & Eye Tracking$\uparrow$ \\
\midrule
\Baseline & - & 24.46$\pm$0.10 & 71.03$\pm$0.27 & 64.10$\pm$0.60 & 27.82$\pm$1.18 & 53.18$\pm$0.28 & 66.90$\pm$2.47 & 1.76$\pm$0.22 & 3.85$\pm$0.31 \\
\NoContrast & 2.96\% & \textbf{23.56$\pm$0.11}$^{*}$ (3.68\%) & \textbf{72.09$\pm$0.17}$^{*}$ (1.50\%) & 64.83$\pm$0.73 (1.15\%) & 28.14$\pm$1.75 (1.16\%) & 53.17$\pm$0.30 (-0.01\%) & 64.67$\pm$1.66$^{*}$ (-3.34\%) & \textbf{1.91$\pm$0.25} (8.34\%) & 4.31$\pm$0.33$^{*}$ (11.92\%) \\
\NoContrastVHead & 0.66\% & 24.33$\pm$0.10$^{*}$ (0.51\%) & 71.67$\pm$0.24$^{*}$ (0.91\%) & 64.86$\pm$0.74 (1.20\%) & 25.47$\pm$1.40$^{*}$ (-8.45\%) & 53.03$\pm$0.31 (-0.27\%) & 66.67$\pm$1.58 (-0.35\%) & 1.76$\pm$0.23 (-0.54\%) & 4.32$\pm$0.33$^{*}$ (12.12\%) \\
CD-Small-20 & 3.55\% & 23.73$\pm$0.14$^{*}$ (2.96\%) & 71.84$\pm$0.19$^{*}$ (1.15\%) & 64.09$\pm$0.66 (-0.01\%) & 29.25$\pm$1.32$^{*}$ (5.15\%) & 53.45$\pm$0.27 (0.50\%) & 69.05$\pm$2.53 (3.21\%) & 1.79$\pm$0.22 (1.59\%) & 4.37$\pm$0.31$^{*}$ (13.29\%) \\
CD-Drop-0.7 & 3.29\% & 24.06$\pm$0.13$^{*}$ (1.65\%) & 71.79$\pm$0.31$^{*}$ (1.08\%) & \textbf{65.19$\pm$0.70} (1.71\%) & 28.91$\pm$1.64$^{*}$ (3.91\%) & 53.62$\pm$0.40 (0.83\%) & 66.80$\pm$1.72 (-0.15\%) & 1.90$\pm$0.35 (7.76\%) & 4.16$\pm$0.44 (7.92\%) \\
CD-Early-500 & \textbf{\textbf{4.90\%}} & 23.73$\pm$0.10$^{*}$ (2.98\%) & 71.72$\pm$0.19$^{*}$ (0.98\%) & 65.10$\pm$0.60$^{*}$ (1.56\%) & \textbf{30.38$\pm$0.65}$^{*}$ (9.19\%) & \textbf{53.80$\pm$0.29}$^{*}$ (1.18\%) & \textbf{70.55$\pm$2.32}$^{*}$ (5.46\%) & 1.79$\pm$0.22 (1.30\%) & \textbf{4.42$\pm$0.32}$^{*}$ (14.64\%) \\
\bottomrule
\end{tabular}
\end{adjustbox}
\caption{\CommonCaption}
\label{tab:SummaryExperiments}
\end{table*}
}

\newcommand{\tableSummaryExperimentsShort}{
\begin{table}[H]
\centering
\begin{adjustbox}{max width=\linewidth}
\begin{tabular}{l|l|l}
\toprule
Name & \mudeltarel$\uparrow$ & Perplexity$\downarrow$ \\
\midrule
\Baseline & - & 24.46$\pm$0.10 \\
\midrule
\NoContrast & 2.96\% & \textbf{23.56$\pm$0.11}$^{*}$ (3.68\%) \\
\NoContrast-Top-k-200 & 3.65\% & 23.65$\pm$0.10$^{*}$ (3.29\%) \\ 
\midrule 
CD-Small-20 & 3.55\% & 23.73$\pm$0.14$^{*}$ (2.96\%) \\
CD-Drop-0.7 & 3.29\% & 24.06$\pm$0.13$^{*}$ (1.65\%) \\
CD-Early-500 & 4.90\% & 23.73$\pm$0.10$^{*}$ (2.98\%) \\
CD-Early-500-Top-k-200 & \textbf{\textbf{5.69\%}} & 23.77$\pm$0.10$^{*}$ (2.80\%) \\
\bottomrule
\end{tabular}

\end{adjustbox}

\caption{Comparison of \CD\ variants (early checkpoint, smaller model, dropout) against non-contrastive baselines, including the best truncation configurations. The best truncation for both regimes is Top-k=200; CD-Early-500-Top-k-200 achieves the highest overall task improvement at unchanged perplexity.}

\label{tab:SummaryExperimentsShort}
\end{table}
}

% --- New Table ---
\newcommand{\tableSummaryExperimentsSmall}{
\begin{table*}[!t]
\centering
\begin{adjustbox}{max width=\linewidth}
\begin{tabular}{l|l|llllllll}
\toprule
Name & \mudeltarel$\uparrow$ & Perplexity$\downarrow$ & BLiMP$\uparrow$ & BLiMP Supp.$\uparrow$ & Entity Tracking$\uparrow$ & EWoK$\uparrow$ & WUG$\uparrow$ & Reading$\uparrow$ & Eye Tracking$\uparrow$ \\
\midrule
\Baseline & - & 24.46$\pm$0.10 & 71.03$\pm$0.27 & 64.10$\pm$0.60 & 27.82$\pm$1.18 & 53.18$\pm$0.28 & 66.90$\pm$2.47 & 1.76$\pm$0.22 & 3.85$\pm$0.31 \\
\NoContrast & 2.96\% & \textbf{23.56$\pm$0.11}$^{*}$ (3.68\%) & \textbf{72.09$\pm$0.17}$^{*}$ (1.50\%) & 64.83$\pm$0.73 (1.15\%) & 28.14$\pm$1.75 (1.16\%) & 53.17$\pm$0.30 (-0.01\%) & 64.67$\pm$1.66$^{*}$ (-3.34\%) & \textbf{1.91$\pm$0.25} (8.34\%) & 4.31$\pm$0.33$^{*}$ (11.92\%) \\
\NoContrastVHead & 0.66\% & 24.33$\pm$0.10$^{*}$ (0.51\%) & 71.67$\pm$0.24$^{*}$ (0.91\%) & 64.86$\pm$0.74 (1.20\%) & 25.47$\pm$1.40$^{*}$ (-8.45\%) & 53.03$\pm$0.31 (-0.27\%) & 66.67$\pm$1.58 (-0.35\%) & 1.76$\pm$0.23 (-0.54\%) & 4.32$\pm$0.33$^{*}$ (12.12\%) \\
CD-Early-500 & \textbf{\textbf{4.90\%}} & 23.73$\pm$0.10$^{*}$ (2.98\%) & 71.72$\pm$0.19$^{*}$ (0.98\%) & \textbf{65.10$\pm$0.60}$^{*}$ (1.56\%) & \textbf{30.38$\pm$0.65}$^{*}$ (9.19\%) & \textbf{53.80$\pm$0.29}$^{*}$ (1.18\%) & \textbf{70.55$\pm$2.32}$^{*}$ (5.46\%) & 1.79$\pm$0.22 (1.30\%) & \textbf{4.42$\pm$0.32}$^{*}$ (14.64\%) \\
\bottomrule
\end{tabular}
\end{adjustbox}
\caption{Task-by-task results for synthetic-data regimes. Entries are mean $\pm$ s.e.; parentheses show relative change vs.\ \Baseline; $^{*}$ denotes a significant difference vs.\ \Baseline. CD-Early-500 attains the best overall \mudeltarel\ (+4.90\%) and leads on BLiMP Supplement, Entity Tracking, EWoK, WUG, and Eye Tracking, while \NoContrast\ yields the lowest Perplexity, the best BLiMP and Reading.}

\label{tab:SummaryExperimentsSmall}
\end{table*}
}

\newcommand{\tableSummaryExperimentsSmallShort}{
\begin{table}[!t]
\centering
\begin{adjustbox}{max width=\linewidth}
\begin{tabular}{l|l|l}
\toprule
Name & \mudeltarel$\uparrow$ & Perplexity$\downarrow$ \\
\midrule
\Baseline & - & 24.46$\pm$0.10 \\
\NoContrast & 2.96\% & \textbf{23.56$\pm$0.11}$^{*}$ (3.68\%) \\
\NoContrastVHead & 0.66\% & 24.33$\pm$0.10$^{*}$ (0.51\%) \\
CD-Early-500 & {{4.90\%}} & 23.73$\pm$0.10$^{*}$ (2.98\%) \\
\bottomrule
\end{tabular}
\end{adjustbox}

\caption{Aggregate results for synthetic-data generation regimes. \mudeltarel\ is the mean relative improvement over \Baseline\ across all non-perplexity tasks; Perplexity shows mean $\pm$ s.e. (lower is better). Asterisks mark significant differences vs.\ \Baseline; percentages in parentheses are relative changes vs.\ \Baseline. CD-Early-500 yields the best overall gains, while \NoContrast attains the lowest perplexity, see \Cref{tab:AllModels} for a full sweep.}

\label{tab:SummaryExperimentsSmallShort}
\end{table}
}

\newcommand{\tableSummaryExperimentsSmallNoRel}{
\begin{table*}[!h]
\centering
\begin{adjustbox}{max width=\linewidth}
\begin{tabular}{l|l|llllllll}
\toprule
Name & \mudeltarel$\uparrow$ & Perplexity$\downarrow$ & BLiMP$\uparrow$ & BLiMP Supp.$\uparrow$ & Entity Tracking$\uparrow$ & EWoK$\uparrow$ & WUG$\uparrow$ & Reading$\uparrow$ & Eye Tracking$\uparrow$ \\
\midrule
Baseline & - & 24.46$\pm$0.10 & 71.03$\pm$0.27 & 64.10$\pm$0.60 & 27.82$\pm$1.18 & 53.18$\pm$0.28 & 66.90$\pm$2.47 & 1.76$\pm$0.22 & 3.85$\pm$0.31 \\
No-Contrast & 2.96\% & \textbf{23.56$\pm$0.11}$^{*}$ & \textbf{72.09$\pm$0.17}$^{*}$ & 64.83$\pm$0.73 & 28.14$\pm$1.75 & 53.17$\pm$0.30 & 64.67$\pm$1.66$^{*}$ & \textbf{1.91$\pm$0.25} & 4.31$\pm$0.33$^{*}$ \\
No-Contrast-V-Head & 0.66\% & 24.33$\pm$0.10$^{*}$ & 71.67$\pm$0.24$^{*}$ & 64.86$\pm$0.74 & 25.47$\pm$1.40$^{*}$ & 53.03$\pm$0.31 & 66.67$\pm$1.58 & 1.76$\pm$0.23 & 4.32$\pm$0.33$^{*}$ \\
CD-Early-500 & \textbf{\textbf{4.90\%}} & 23.73$\pm$0.10$^{*}$ & 71.72$\pm$0.19$^{*}$ & \textbf{65.10$\pm$0.60}$^{*}$ & \textbf{30.38$\pm$0.65}$^{*}$ & \textbf{53.80$\pm$0.29}$^{*}$ & \textbf{70.55$\pm$2.32}$^{*}$ & 1.79$\pm$0.22 & \textbf{4.42$\pm$0.32}$^{*}$ \\
\bottomrule
\end{tabular}
\end{adjustbox}

\caption{Task-by-task results for synthetic-data regimes. Entries are mean $\pm$ s.e.; $^{*}$ denotes a significant difference vs.\ \Baseline. CD-Early-500 attains the best overall \mudeltarel\ (+4.90\%) and leads on BLiMP Supplement, Entity Tracking, EWoK, WUG, and Eye Tracking, while \NoContrast\ yields the lowest Perplexity, the best BLiMP and Reading. Find relative change vs.\ \Baseline at \Cref{tab:AllModels}}

\label{tab:SummaryExperimentsSmallNoRel}
\end{table*}
}

\newcommand{\tableCDEarlyVsNoContrastDeltaTransposedSign}{
\begin{table}[H]
\centering
\begin{adjustbox}{max width=\linewidth}
\begin{tabular}{l r l}
\toprule
\textbf{Metric} & \textsc{CD} vs \NoContrast & Significance \\
\midrule
${\mudeltarel}_{\CD} - {\mudeltarel}_{\NoContrast}$ & +1.94pp &  \\
Perplexity$\downarrow$ & -0.7\% & $^{***}$ \\
BLiMP$\uparrow$ & -0.5\% & $^{***}$ \\
BLiMP Supp.$\uparrow$ & +0.4\% & \\
Entity Tracking$\uparrow$ & +7.3\% & $^{***}$ \\
EWoK$\uparrow$ & +1.2\% & $^*$ \\
WUG$\uparrow$ & +8.2\% & $^{***}$ \\
Reading$\uparrow$ & -6.2\% & \\
Eye Tracking$\uparrow$ & +2.5\% & \\
\bottomrule
\end{tabular}
\end{adjustbox}

\caption{Statistical significance and relative change of \textsc{CD--Early--500} vs.\ \NoContrast{} by metric. Entries are percentage changes; for Perplexity ($\downarrow$), more negative is better, while for all others ($\uparrow$), more positive is better. The “Significance” column reports paired-bootstrap one-sided $p$-values per \Cref{sec:eval_statistics}: $^*$ $p{<}0.05$, $^{**}$ $p{<}0.01$, $^{***}$ $p{<}0.001$ (blank = not significant). \mudeltarel{} is shown as an absolute difference in percentage points (pp).}

\label{tab:CDEarlyVsNoContrastDeltaTransposed}
\end{table}
}

% --- New Table ---
\newcommand{\tableSecondGen}{
\begin{table*}[!t]
\centering
\begin{adjustbox}{max width=\linewidth}
\begin{tabular}{l|l|llllllll}
\toprule
Name & \mudeltarel$\uparrow$ & Perplexity$\downarrow$ & BLiMP$\uparrow$ & BLiMP Supp.$\uparrow$ & Entity Tracking$\uparrow$ & EWoK$\uparrow$ & WUG$\uparrow$ & Reading$\uparrow$ & Eye Tracking$\uparrow$ \\
\midrule
Baseline & - & 24.46$\pm$0.10 & 71.03$\pm$0.27 & 64.10$\pm$0.60 & 27.82$\pm$1.18 & 53.18$\pm$0.28 & 66.90$\pm$2.47 & 1.76$\pm$0.22 & 3.85$\pm$0.31 \\
sec\_gen\_g5000\_1000\_cs1\_mr03 & 3.89\% & 23.83$\pm$0.19$^{*}$ (2.57\%) & 71.87$\pm$0.13$^{*}$ (1.18\%) & 64.25$\pm$1.11 (0.23\%) & 30.80$\pm$1.59$^{*}$ (10.73\%) & 53.71$\pm$0.43 (1.01\%) & 65.25$\pm$3.12 (-2.47\%) & 1.86$\pm$0.37 (5.52\%) & 4.28$\pm$0.48 (11.02\%) \\
sec\_gen\_g5000\_1500\_cs1\_mr03 & 3.40\% & 23.86$\pm$0.16$^{*}$ (2.46\%) & 71.86$\pm$0.25$^{*}$ (1.17\%) & 64.22$\pm$0.79 (0.19\%) & 28.68$\pm$1.57$^{*}$ (3.10\%) & 53.62$\pm$0.41 (0.83\%) & 67.10$\pm$2.18 (0.30\%) & \textbf{1.89$\pm$0.32} (7.31\%) & 4.27$\pm$0.43 (10.87\%) \\
sec\_gen\_g5000\_2000\_cs1\_mr03 & \textbf{\textbf{4.12\%}} & \textbf{23.71$\pm$0.13}$^{*}$ (3.04\%) & \textbf{72.05$\pm$0.27}$^{*}$ (1.44\%) & \textbf{64.98$\pm$0.94} (1.38\%) & 30.37$\pm$1.55$^{*}$ (9.17\%) & 53.13$\pm$0.38 (-0.08\%) & \textbf{69.80$\pm$3.80}$^{*}$ (4.33\%) & 1.87$\pm$0.34 (5.95\%) & 4.11$\pm$0.42 (6.67\%) \\
sec\_gen\_g5000\_500\_cs1\_mr03 & 3.63\% & 23.75$\pm$0.12$^{*}$ (2.91\%) & 71.85$\pm$0.23$^{*}$ (1.16\%) & 64.50$\pm$0.62 (0.63\%) & \textbf{31.21$\pm$1.22}$^{*}$ (12.18\%) & \textbf{53.81$\pm$0.31}$^{*}$ (1.19\%) & 65.44$\pm$1.49 (-2.18\%) & 1.80$\pm$0.23 (1.98\%) & 4.26$\pm$0.33$^{*}$ (10.45\%) \\
\bottomrule
\end{tabular}
\end{adjustbox}
\caption{Second Gen Models. $^{*}$ marks statistical significance; percentages indicate relative change vs Baseline.}
\label{tab:SecondGen}
\end{table*}
}

\newcommand{\tableSecondGenShort}{
\begin{table}[]
\centering
\begin{adjustbox}{max width=\linewidth}
\begin{tabular}{l|l|l}
\toprule
Name & \mudeltarel$\uparrow$ & Perplexity$\downarrow$ \\
\midrule
Baseline & - & 24.46$\pm$0.10 \\
sec\_gen\_g5000\_1000\_cs1\_mr03 & 3.89\% & 23.83$\pm$0.19$^{*}$ (2.57\%) \\
sec\_gen\_g5000\_1500\_cs1\_mr03 & 3.40\% & 23.86$\pm$0.16$^{*}$ (2.46\%) \\
sec\_gen\_g5000\_2000\_cs1\_mr03 & \textbf{\textbf{4.12\%}} & \textbf{23.71$\pm$0.13}$^{*}$ (3.04\%) \\
sec\_gen\_g5000\_500\_cs1\_mr03 & 3.63\% & 23.75$\pm$0.12$^{*}$ (2.91\%) \\
\bottomrule
\end{tabular}
\end{adjustbox}
\caption{Second Gen Models. $^{*}$ marks statistical significance; percentages indicate relative change vs Baseline.}
\label{tab:SecondGenShort}
\end{table}
}

% --- New Table ---
\newcommand{\tableTopKP}{
\begin{table*}[!h]
\centering
\begin{adjustbox}{max width=\linewidth}
\begin{tabular}{l|l|llllllll}
\toprule
Name & \mudeltarel$\uparrow$ & Perplexity$\downarrow$ & BLiMP$\uparrow$ & BLiMP Supp.$\uparrow$ & Entity Tracking$\uparrow$ & EWoK$\uparrow$ & WUG$\uparrow$ & Reading$\uparrow$ & Eye Tracking$\uparrow$ \\
\midrule
Baseline & - & 24.46$\pm$0.10 & 71.03$\pm$0.27 & 64.10$\pm$0.60 & 27.82$\pm$1.18 & 53.18$\pm$0.28 & 66.90$\pm$2.47 & 1.76$\pm$0.22 & 3.85$\pm$0.31 \\
\midrule
No-Contrast & 2.96\% & \textbf{23.56$\pm$0.11}$^{*}$ (3.68\%) & 72.09$\pm$0.17$^{*}$ (1.50\%) & 64.83$\pm$0.73 (1.15\%) & 28.14$\pm$1.75 (1.16\%) & 53.17$\pm$0.30 (-0.01\%) & 64.67$\pm$1.66$^{*}$ (-3.34\%) & 1.91$\pm$0.25 (8.34\%) & 4.31$\pm$0.33$^{*}$ (11.92\%) \\
No-Contrast-V-Head & 0.66\% & 24.33$\pm$0.10$^{*}$ (0.51\%) & 71.67$\pm$0.24$^{*}$ (0.91\%) & 64.86$\pm$0.74 (1.20\%) & 25.47$\pm$1.40$^{*}$ (-8.45\%) & 53.03$\pm$0.31 (-0.27\%) & 66.67$\pm$1.58 (-0.35\%) & 1.76$\pm$0.23 (-0.54\%) & 4.32$\pm$0.33$^{*}$ (12.12\%) \\
\midrule
No-Contrast-TopK-50 & 3.48\% & 23.88$\pm$0.10$^{*}$ (2.36\%) & 71.52$\pm$0.13$^{*}$ (0.69\%) & 64.45$\pm$0.83 (0.54\%) & 27.23$\pm$1.64$^{*}$ (-2.13\%) & 53.37$\pm$0.30 (0.37\%) & 67.38$\pm$1.82 (0.71\%) & \textbf{1.97$\pm$0.27} (11.83\%) & 4.33$\pm$0.36$^{*}$ (12.38\%) \\
No-Contrast-TopK-100 & 2.49\% & 23.81$\pm$0.11$^{*}$ (2.66\%) & 72.12$\pm$0.26$^{*}$ (1.55\%) & 64.22$\pm$0.69 (0.19\%) & 28.09$\pm$1.65 (0.98\%) & 53.43$\pm$0.32 (0.48\%) & 66.71$\pm$2.15 (-0.28\%) & 1.85$\pm$0.27 (4.65\%) & 4.23$\pm$0.38 (9.86\%) \\
No-Contrast-TopK-200 & 3.65\% & 23.65$\pm$0.10$^{*}$ (3.29\%) & 71.78$\pm$0.21$^{*}$ (1.06\%) & 63.98$\pm$0.69 (-0.19\%) & 26.96$\pm$1.23$^{*}$ (-3.08\%) & 53.52$\pm$0.32 (0.64\%) & 67.81$\pm$1.53 (1.36\%) & 1.96$\pm$0.26 (10.76\%) & 4.43$\pm$0.35$^{*}$ (15.01\%) \\
No-Contrast-TopP-90 & 2.73\% & 23.88$\pm$0.11$^{*}$ (2.37\%) & 71.96$\pm$0.14$^{*}$ (1.31\%) & 64.84$\pm$0.62 (1.16\%) & 26.12$\pm$0.89$^{*}$ (-6.09\%) & 53.36$\pm$0.27 (0.35\%) & 66.25$\pm$2.05 (-0.97\%) & 1.94$\pm$0.23 (9.92\%) & 4.37$\pm$0.32$^{*}$ (13.44\%) \\
No-Contrast-TopP-95 & 2.33\% & 23.74$\pm$0.12$^{*}$ (2.93\%) & 72.02$\pm$0.22$^{*}$ (1.40\%) & 64.50$\pm$0.63 (0.63\%) & 26.29$\pm$1.31$^{*}$ (-5.51\%) & 53.41$\pm$0.32 (0.45\%) & 66.44$\pm$1.52 (-0.69\%) & 1.90$\pm$0.26 (7.51\%) & 4.33$\pm$0.35$^{*}$ (12.51\%) \\
No-Contrast-TopP-97 & 2.11\% & 23.61$\pm$0.10$^{*}$ (3.47\%) & 71.62$\pm$0.11$^{*}$ (0.83\%) & 64.33$\pm$0.68 (0.36\%) & 27.29$\pm$1.48$^{*}$ (-1.91\%) & 53.24$\pm$0.28 (0.12\%) & 66.00$\pm$1.63 (-1.35\%) & 1.87$\pm$0.24 (6.20\%) & 4.26$\pm$0.32$^{*}$ (10.51\%) \\
\midrule
CD-Early-500 & 4.90\% & 23.73$\pm$0.10$^{*}$ (2.98\%) & 71.72$\pm$0.19$^{*}$ (0.98\%) & 65.10$\pm$0.60$^{*}$ (1.56\%) & 30.38$\pm$0.65$^{*}$ (9.19\%) & 53.80$\pm$0.29$^{*}$ (1.18\%) & \textbf{70.55$\pm$2.32}$^{*}$ (5.46\%) & 1.79$\pm$0.22 (1.30\%) & 4.42$\pm$0.32$^{*}$ (14.64\%) \\
\midrule
CD-Early-500-TopK-50 & 4.64\% & 23.90$\pm$0.12$^{*}$ (2.30\%) & 71.90$\pm$0.21$^{*}$ (1.23\%) & 64.74$\pm$0.68 (1.01\%) & 30.29$\pm$1.49$^{*}$ (8.89\%) & 53.47$\pm$0.32 (0.55\%) & 67.56$\pm$2.48 (0.99\%) & 1.93$\pm$0.26 (9.49\%) & 4.25$\pm$0.35 (10.30\%) \\
CD-Early-500-TopK-100 & 4.90\% & 23.79$\pm$0.12$^{*}$ (2.73\%) & 71.49$\pm$0.18$^{*}$ (0.65\%) & \textbf{65.29$\pm$0.80}$^{*}$ (1.87\%) & \textbf{33.11$\pm$0.62}$^{*}$ (19.02\%) & \textbf{53.94$\pm$0.36}$^{*}$ (1.44\%) & 67.44$\pm$2.13 (0.80\%) & 1.73$\pm$0.26 (-2.20\%) & 4.34$\pm$0.35$^{*}$ (12.74\%) \\
CD-Early-500-TopK-200 & \textbf{\textbf{5.69\%}} & 23.77$\pm$0.10$^{*}$ (2.80\%) & 71.87$\pm$0.35$^{*}$ (1.19\%) & 64.23$\pm$0.59 (0.20\%) & 31.05$\pm$0.79$^{*}$ (11.61\%) & 53.61$\pm$0.31 (0.82\%) & 67.90$\pm$2.00 (1.49\%) & 1.92$\pm$0.25 (8.73\%) & \textbf{4.46$\pm$0.33}$^{*}$ (15.78\%) \\
CD-Early-500-TopP-90 & 4.91\% & 23.74$\pm$0.10$^{*}$ (2.93\%) & \textbf{72.16$\pm$0.14}$^{*}$ (1.60\%) & 64.69$\pm$0.65 (0.92\%) & 30.43$\pm$1.07$^{*}$ (9.37\%) & 53.68$\pm$0.30 (0.94\%) & 68.35$\pm$1.44 (2.17\%) & 1.85$\pm$0.23 (4.59\%) & 4.42$\pm$0.33$^{*}$ (14.77\%) \\
CD-Early-500-TopP-95 & 4.54\% & 23.80$\pm$0.15$^{*}$ (2.69\%) & 71.36$\pm$0.27$^{*}$ (0.47\%) & 64.79$\pm$0.62 (1.09\%) & 32.56$\pm$0.74$^{*}$ (17.06\%) & 53.60$\pm$0.29 (0.80\%) & 65.78$\pm$1.99 (-1.68\%) & 1.82$\pm$0.24 (2.86\%) & 4.28$\pm$0.33 (11.14\%) \\
CD-Early-500-TopP-97 & 2.98\% & 23.86$\pm$0.13$^{*}$ (2.46\%) & 71.69$\pm$0.20$^{*}$ (0.94\%) & 64.54$\pm$0.56 (0.69\%) & 30.20$\pm$0.92$^{*}$ (8.56\%) & 53.63$\pm$0.30 (0.86\%) & 64.85$\pm$1.50 (-3.06\%) & 1.82$\pm$0.22 (3.06\%) & 4.23$\pm$0.31 (9.84\%) \\
\bottomrule
\end{tabular}
\end{adjustbox}
\caption{Top K and Top P Experiment. $^{*}$ marks statistical significance; percentages indicate relative change vs Baseline.}
\label{tab:TopKP}
\end{table*}
}

\newcommand{\tableTopKPShort}{
\begin{table}[!h]
\centering
\begin{adjustbox}{max width=\linewidth}
\begin{tabular}{l|l|l}
\toprule
Name & \mudeltarel$\uparrow$ & Perplexity$\downarrow$ \\
\midrule
Baseline & - & 24.46$\pm$0.10 \\
No-Contrast & 2.96\% & \textbf{23.56$\pm$0.11}$^{*}$ (3.68\%) \\
No-Contrast-V-Head & 0.66\% & 24.33$\pm$0.10$^{*}$ (0.51\%) \\
No-Contrast-TopK-50 & 3.48\% & 23.88$\pm$0.10$^{*}$ (2.36\%) \\
No-Contrast-TopK-100 & 2.49\% & 23.81$\pm$0.11$^{*}$ (2.66\%) \\
No-Contrast-TopK-200 & 3.65\% & 23.65$\pm$0.10$^{*}$ (3.29\%) \\
No-Contrast-TopP-90 & 2.73\% & 23.88$\pm$0.11$^{*}$ (2.37\%) \\
No-Contrast-TopP-95 & 2.33\% & 23.74$\pm$0.12$^{*}$ (2.93\%) \\
No-Contrast-TopP-97 & 2.11\% & 23.61$\pm$0.10$^{*}$ (3.47\%) \\
CD-Early-500 & 4.90\% & 23.73$\pm$0.10$^{*}$ (2.98\%) \\
CD-Early-500-TopK-50 & 4.64\% & 23.90$\pm$0.12$^{*}$ (2.30\%) \\
CD-Early-500-TopK-100 & 4.90\% & 23.79$\pm$0.12$^{*}$ (2.73\%) \\
CD-Early-500-TopK-200 & \textbf{\textbf{5.69\%}} & 23.77$\pm$0.10$^{*}$ (2.80\%) \\
CD-Early-500-TopP-90 & 4.91\% & 23.74$\pm$0.10$^{*}$ (2.93\%) \\
CD-Early-500-TopP-95 & 4.54\% & 23.80$\pm$0.15$^{*}$ (2.69\%) \\
CD-Early-500-TopP-97 & 2.98\% & 23.86$\pm$0.13$^{*}$ (2.46\%) \\
\bottomrule
\end{tabular}
\end{adjustbox}
\caption{Top K and Top P Experiment. $^{*}$ marks statistical significance; percentages indicate relative change vs Baseline.}
\label{tab:TopKPShort}
\end{table}
}

\newcommand{\tableTopKPNoRel}{
\begin{table*}[!h]
\centering
\begin{adjustbox}{max width=\linewidth}
\begin{tabular}{l|l|llllllll}
\toprule
Name & \mudeltarel$\uparrow$ & Perplexity$\downarrow$ & BLiMP$\uparrow$ & BLiMP Supp.$\uparrow$ & Entity Tracking$\uparrow$ & EWoK$\uparrow$ & WUG$\uparrow$ & Reading$\uparrow$ & Eye Tracking$\uparrow$ \\
\midrule
Baseline & - & 24.46$\pm$0.10 & 71.03$\pm$0.27 & 64.10$\pm$0.60 & 27.82$\pm$1.18 & 53.18$\pm$0.28 & 66.90$\pm$2.47 & 1.76$\pm$0.22 & 3.85$\pm$0.31 \\
No-Contrast & 2.96\% & \textbf{23.56$\pm$0.11}$^{*}$ & 72.09$\pm$0.17$^{*}$ & 64.83$\pm$0.73 & 28.14$\pm$1.75 & 53.17$\pm$0.30 & 64.67$\pm$1.66$^{*}$ & 1.91$\pm$0.25 & 4.31$\pm$0.33$^{*}$ \\
No-Contrast-V-Head & 0.66\% & 24.33$\pm$0.10$^{*}$ & 71.67$\pm$0.24$^{*}$ & 64.86$\pm$0.74 & 25.47$\pm$1.40$^{*}$ & 53.03$\pm$0.31 & 66.67$\pm$1.58 & 1.76$\pm$0.23 & 4.32$\pm$0.33$^{*}$ \\
No-Contrast-TopK-50 & 3.48\% & 23.88$\pm$0.10$^{*}$ & 71.52$\pm$0.13$^{*}$ & 64.45$\pm$0.83 & 27.23$\pm$1.64$^{*}$ & 53.37$\pm$0.30 & 67.38$\pm$1.82 & \textbf{1.97$\pm$0.27} & 4.33$\pm$0.36$^{*}$ \\
No-Contrast-TopK-100 & 2.49\% & 23.81$\pm$0.11$^{*}$ & 72.12$\pm$0.26$^{*}$ & 64.22$\pm$0.69 & 28.09$\pm$1.65 & 53.43$\pm$0.32 & 66.71$\pm$2.15 & 1.85$\pm$0.27 & 4.23$\pm$0.38 \\
No-Contrast-TopK-200 & 3.65\% & 23.65$\pm$0.10$^{*}$ & 71.78$\pm$0.21$^{*}$ & 63.98$\pm$0.69 & 26.96$\pm$1.23$^{*}$ & 53.52$\pm$0.32 & 67.81$\pm$1.53 & 1.96$\pm$0.26 & 4.43$\pm$0.35$^{*}$ \\
No-Contrast-TopP-90 & 2.73\% & 23.88$\pm$0.11$^{*}$ & 71.96$\pm$0.14$^{*}$ & 64.84$\pm$0.62 & 26.12$\pm$0.89$^{*}$ & 53.36$\pm$0.27 & 66.25$\pm$2.05 & 1.94$\pm$0.23 & 4.37$\pm$0.32$^{*}$ \\
No-Contrast-TopP-95 & 2.33\% & 23.74$\pm$0.12$^{*}$ & 72.02$\pm$0.22$^{*}$ & 64.50$\pm$0.63 & 26.29$\pm$1.31$^{*}$ & 53.41$\pm$0.32 & 66.44$\pm$1.52 & 1.90$\pm$0.26 & 4.33$\pm$0.35$^{*}$ \\
No-Contrast-TopP-97 & 2.11\% & 23.61$\pm$0.10$^{*}$ & 71.62$\pm$0.11$^{*}$ & 64.33$\pm$0.68 & 27.29$\pm$1.48$^{*}$ & 53.24$\pm$0.28 & 66.00$\pm$1.63 & 1.87$\pm$0.24 & 4.26$\pm$0.32$^{*}$ \\
CD-Early-500 & 4.90\% & 23.73$\pm$0.10$^{*}$ & 71.72$\pm$0.19$^{*}$ & 65.10$\pm$0.60$^{*}$ & 30.38$\pm$0.65$^{*}$ & 53.80$\pm$0.29$^{*}$ & \textbf{70.55$\pm$2.32}$^{*}$ & 1.79$\pm$0.22 & 4.42$\pm$0.32$^{*}$ \\
CD-Early-500-TopK-50 & 4.64\% & 23.90$\pm$0.12$^{*}$ & 71.90$\pm$0.21$^{*}$ & 64.74$\pm$0.68 & 30.29$\pm$1.49$^{*}$ & 53.47$\pm$0.32 & 67.56$\pm$2.48 & 1.93$\pm$0.26 & 4.25$\pm$0.35 \\
CD-Early-500-TopK-100 & 4.90\% & 23.79$\pm$0.12$^{*}$ & 71.49$\pm$0.18$^{*}$ & \textbf{65.29$\pm$0.80}$^{*}$ & \textbf{33.11$\pm$0.62}$^{*}$ & \textbf{53.94$\pm$0.36}$^{*}$ & 67.44$\pm$2.13 & 1.73$\pm$0.26 & 4.34$\pm$0.35$^{*}$ \\
CD-Early-500-TopK-200 & \textbf{\textbf{5.69\%}} & 23.77$\pm$0.10$^{*}$ & 71.87$\pm$0.35$^{*}$ & 64.23$\pm$0.59 & 31.05$\pm$0.79$^{*}$ & 53.61$\pm$0.31 & 67.90$\pm$2.00 & 1.92$\pm$0.25 & \textbf{4.46$\pm$0.33}$^{*}$ \\
CD-Early-500-TopP-90 & 4.91\% & 23.74$\pm$0.10$^{*}$ & \textbf{72.16$\pm$0.14}$^{*}$ & 64.69$\pm$0.65 & 30.43$\pm$1.07$^{*}$ & 53.68$\pm$0.30 & 68.35$\pm$1.44 & 1.85$\pm$0.23 & 4.42$\pm$0.33$^{*}$ \\
CD-Early-500-TopP-95 & 4.54\% & 23.80$\pm$0.15$^{*}$ & 71.36$\pm$0.27$^{*}$ & 64.79$\pm$0.62 & 32.56$\pm$0.74$^{*}$ & 53.60$\pm$0.29 & 65.78$\pm$1.99 & 1.82$\pm$0.24 & 4.28$\pm$0.33 \\
CD-Early-500-TopP-97 & 2.98\% & 23.86$\pm$0.13$^{*}$ & 71.69$\pm$0.20$^{*}$ & 64.54$\pm$0.56 & 30.20$\pm$0.92$^{*}$ & 53.63$\pm$0.30 & 64.85$\pm$1.50 & 1.82$\pm$0.22 & 4.23$\pm$0.31 \\
\bottomrule
\end{tabular}
\end{adjustbox}
\caption{Top K and Top P Experiment. $^{*}$ marks statistical significance; percentages indicate relative change vs Baseline.}
\label{tab:TopKPNoRel}
\end{table*}
}

\section{Introduction}
\label{sec:intro}

\begin{figure}[!t]
    \centering
    \includegraphics[width=0.8\linewidth]{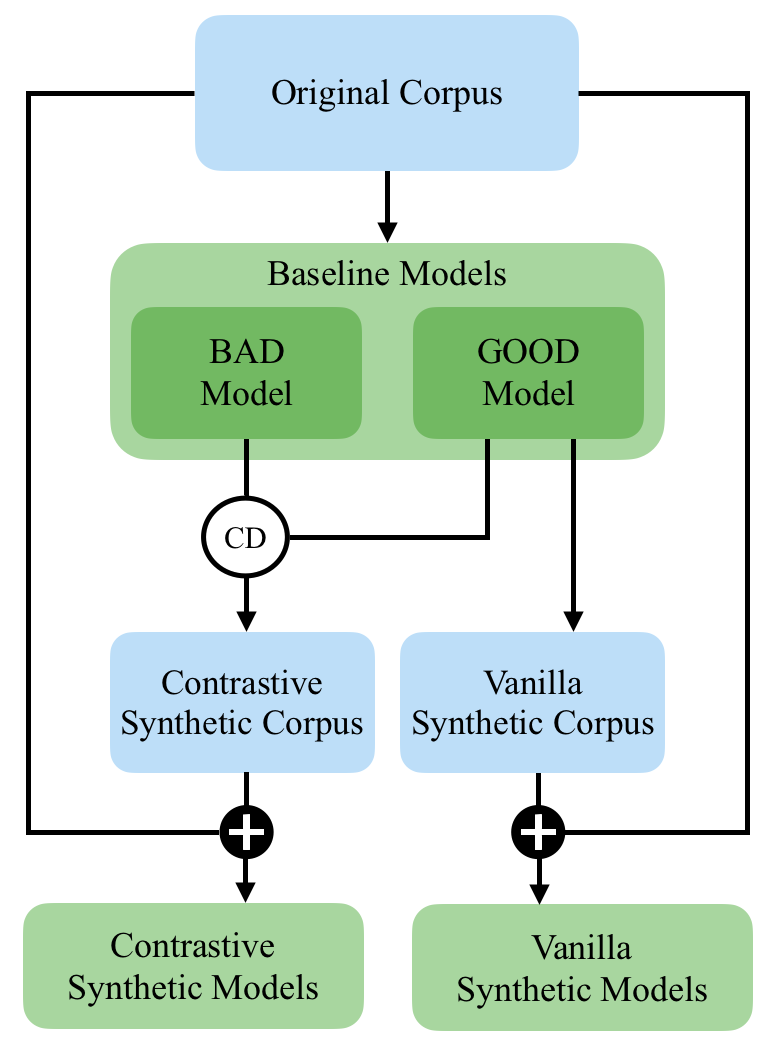}
    \caption{Our synthetic data generation and training pipeline: Start by training baseline LMs on a ``real'' corpus (\emph{TinyBabyLM}: human-written text + \emph{TinyStories}). The \good model is the best checkpoint; the \bad model is a weaker variant, e.g., an earlier checkpoint. We generate synthetic corpora via (i) \emph{contrastive decoding (CD)}, and (ii) non-contrastive ancestral (\emph{vanilla}) sampling. We then train new models on a mixture of the original and synthetic corpora. We find that contrastive models improve the most over the \Baseline in evaluations on reasoning-oriented benchmarks, such as entity tracking. \looseness=-1 
    }
    \label{fig:dataflow}
\end{figure}

\looseness=-1

Large language models (LLMs) require enormous amounts of text to achieve strong performance \cite{kaplan_scaling_2020,hoffmann_training_2022}.
For the largest models, it has even been claimed that current training regimes already consume the vast majority of publicly available text on the internet \cite{villalobos_will_2024,grattafiori_llama_2024}.
The BabyLM Challenge \cite{charpentier_babylm_2025} emphasizes this point by asking what can be learned under a strict budget of 100M words, prioritizing data efficiency over raw scale, mimicking the far more efficient language learning capabilities of humans.
Furthermore, not all training data is equally beneficial \cite{eldan_tinystories_2023, gunasekar_textbooks_2023}.
The question thus arises: How can we get more high-quality data in a constrained setting?
One proposed solution is to generate synthetic data using existing pre-trained models, thereby expanding the available corpus without collecting more human-written text \cite{wang_self-instruct_2023,eldan_tinystories_2023,gunasekar_textbooks_2023, abdin_phi-4_2024}.\looseness=-1

Generating synthetic data is non-trivial, however. The quality of synthetic text may be hindered by noise, factual errors, or stylistic artifacts \cite{lin_truthfulqa_2022,huang_survey_2025}. Models may also replicate or even amplify biases from their training data \cite{gallegos_bias_2024, bender_dangers_2021}, and generated text may diverge from the target distribution, leading to potential degradation in downstream performance or model collapse \cite{dohmatob_strong_2024, gerstgrasser_is_2024, shumailov_ai_2024}.
Moreover, producing high-quality synthetic data is particularly difficult because language models often hallucinate facts or repeat memorized content from their original training corpus \cite{bender_dangers_2021,lin_truthfulqa_2022}.

This work explores the use of contrastive decoding ($\CD$) \cite{li_contrastive_2023} to generate synthetic data in a controlled setting. $\CD$ is a decoding strategy that takes advantage of the differences between a \good model and a \bad model to produce more coherent and informative text.
In prior work, $\CD$ has been largely restricted to improving the quality of responses generated for inference-time tasks \cite{li_contrastive_2023,obrien_contrastive_2023,chang_explaining_2024}.
In contrast, we use $\CD$ to synthesize corpora to train new models from scratch. Our goal is to know whether these inference-time benefits of $\CD$ translate into gains when generating synthetic data for training language models.\looseness=-1

The high-level experimental approach is illustrated in \Cref{fig:dataflow} and goes as follows.
\begin{enumerate}
    \item Start with an original corpus (100M tokens, BabyLM setting \cite{charpentier_babylm_2025}).
	\item Train \Baseline models (100M-parameter models based on the Llama 2 architecture \citep{touvron_llama_2023}) on the original corpus.
	\item Generate synthetic corpora (100M tokens each) using $\CD$ and standard sampling.
	\item Train models on the original and synthetic corpora.
    \item Evaluate models on downstream tasks and compare to \Baseline.
\end{enumerate}

We find that synthetic data improves performance on the language-modeling objective and downstream tasks. Moreover, tasks that emphasize reasoning benefit most from $\CD$-generated data, whereas tasks emphasizing linguistic competence gain more from standard (non-contrastive) sampling.\looseness=-1

\section{Synthetic Data Generation for Pre-training Language Models}

Recent work shows that \emph{curated, high-quality} synthetic corpora can substantially boost data efficiency for small or low-resource LMs \cite{eldan_tinystories_2023}. Carefully constructed ``textbook''-style corpora improve generalization \cite{gunasekar_textbooks_2023}, and iterative pipelines that generate, critique, and revise synthetic content have been shown to boost reasoning-oriented capabilities \cite{abdin_phi-4_2024}.
Domain-targeted corpora can be especially effective: \textit{TinyStories} demonstrates that fully synthetic, child-directed narratives enable $1$--$10$M-parameter models to produce multi-paragraph coherent and grammatical text \cite{eldan_tinystories_2023}.
For instruction following, Self-Instruct bootstraps instruction-response pairs from a seed set, leading to gains without additional human annotation \cite{wang_self-instruct_2023}. These results collectively suggest that synthetic data can significantly increase downstream performance.\looseness=-1

However, naive reuse of model-generated text across generations can severly harm performance, resulting in ``model collapse'' \cite{shumailov_ai_2024,gerstgrasser_is_2024,dohmatob_strong_2024}. 
Empirically, careful filtering, diversification, and sustained mixing with real data mitigate such risks while preserving gains \cite{gerstgrasser_is_2024}. In this work, we explore an orthogonal axis: \emph{decoding-control} for synthetic corpora. Specifically, we study whether $\CD$ can produce higher-signal synthetic corpora for pre-training under a tight data budget, compared to non contrastive approaches.\looseness=-1

\section{Contrastive Decoding}

\paragraph{Language-models.}
Following \citet{cotterell_formal_2024}, let $\alphabet$ be a set of tokens we call the vocabulary, the Kleene closure $\alphabet^*$ be the set of all strings built from $\alphabet$, if $p$ is a probability distribution over $\Sigma^*$ we say it is a \textbf{language model}.
Then, $p(\x \mid \xstr)$ represents the model’s \emph{next-token} probability, i.e., the probability that the next token is $\x$ given the preceding context $\xstr \defeq x_0x_1 \ldots x_{i-1}$. 

\paragraph{Contrastive Decoding}
We now describe the $\CD$ approach in detail.
Let $\pexp$ be a \good (better performing) language model, and $\pama$ be a \bad (worse performing) language model.
Following \citet{li_contrastive_2023}, we define $\vhead$ as the set of likely tokens under $\pexp$:
\begin{align}   
\vhead&\left(\xstr\right)\defeq \{\x \in \alphabet: \label{eq:VHEAD}\\
&\vspace{-1em}\pexp\left(\x \mid \xstr\right) \geq \alpha \max_{w \in \alphabet} \pexp\left(w \mid \xstr\right) \}.\notag
\end{align}   
Where $\alpha$ is a scalar hyper-parameter.
The contrastive score $\CDsymb$ for $\x \in \vhead\left(\xstr\right)$ is then defined as follows:
\begin{align}   
\CDsymb&\left(\x \mid \xstr\right) 
\defeq \label{eq:CD_score} \\ &\log \pexp\left(\x \mid \xstr\right) -\lambda  \log \pama\left(\x \mid \xstr\right)\notag,
\end{align}
where contrast strength is controlled by a scalar $\lambda$.
Further, if $\x \notin \vhead\left(\xstr\right)$ then $\CDsymb \left(\x \mid \xstr\right) \defeq-\inf$.
Typically, the contrastive scores $\CDsymb\left(\cdot \mid \xstr\right)$ are treated as logits giving rise to a new probability distribution over $\alphabet$ from which we can decode the next token.

\paragraph{Background and variants.}
$\CD$ biases generation toward tokens preferred by a stronger \good model while down-weighting those preferred by a weaker \bad model, under the plausibility mask $\vhead$ \cite{li_contrastive_2023}. Empirically, $\CD$ reduces repetition and topic drift in open-ended generation and, without additional training, improves reasoning-focused decoding compared to greedy or nucleus (top-$p$) sampling \cite{li_contrastive_2023,obrien_contrastive_2023}. 

Several works adapt $\CD$ to lower its compute and memory cost or to strengthen specific capabilities. \citet{phan_distillation_2024} replace an explicit bad model with a distilled proxy (e.g., via dropout or quantization), retaining most of $\CD$’s gains while reducing memory. In retrieval or context-heavy settings, \citet{zhao_enhancing_2024} integrate $\CD$ with adversarial negatives so that decoding remains grounded in relevant passages. 
These methods focus on evaluating the $\CD$-like inference performance, rather than on generating pre-training corpora.

\paragraph{Relation to synthetic-data generation.}
A related approach is STEER, which performs contrastive expert guidance by subtracting a base model from a fine-tuned domain expert and combining it with negative prompting to generate synthetic corpora for downstream fine-tuning \cite{oneill_steering_2023}.
In contrast, we use $\CD$ with a general \good/\bad pair trained on the same base corpus and treat $\CD$ as a data generator for pre-training: we synthesize full corpora and then train new models from scratch on mixtures of real and synthetic text. This lets us test whether $\CD$’s inference-time benefits translate into better pre-training signals, and how they compare to vanilla sampling under a fixed data budget.

\section{Training on Synthetic Data}
Given the success of $\CD$ in generating higher scoring text for evaluations, we ask whether it can also be employed to generate higher-quality text for pre-training.
This section describes our procedure for generating synthetic corpora using $\CD$ and training models on them. 
\subsection{Synthetic Corpus Generation}

\paragraph{General Procedure.}
To ensure independence from the training data, following \citep{wang_self-instruct_2023} we generate synthetic corpora from \emph{prefix seeds} that are held out from all training and evaluation data.
The prefix seeds are evenly sampled across the four data sources to preserve balance, we describe this in more detail in \Cref{sec:dataset}. 
For each prefix seed, we fix the first 20 tokens as a context prefix, and then we sample continuations from the target model. 
To ensure sufficient diversity and corpus size, we produce eight completions of up to 400 tokens per seed. 
To sample each next token, we use the decoding strategies described below.
Using $\sim$30.4K generation seeds we produce approximately $100$M tokens for each decoding strategy. 

\paragraph{Decoding Strategies.}
\label{sec:decoding_strategies}
We mainly compare two decoding settings that differ only in how candidate tokens are scored before sampling. Let $\vhead(\xstr)$ be the set of $\alpha$-likely tokens of the \good distribution $\pexp$ as defined in Eq.~\eqref{eq:VHEAD}; If $\vhead(\xstr)$ is applied, tokens outside $\vhead$ are assigned score $-\infty$ \cite{li_contrastive_2023}. Let the contrastive score $\CDsymb(\x \mid \xstr)$ be as in Eq.~\eqref{eq:CD_score}. For $\CD$ we treat $\CDsymb(\cdot\mid \xstr)$ as a logit over $\vhead(\xstr)$, i.e., we sample with probabilities proportional to $\exp\big(\CD(\x\mid \xstr)\big)$.

\begin{enumerate}
    \item \textbf{\NoContrast:} Ancestral sampling from $\pexp\left(\cdot\mid \xstr\right)$.
    \item \textbf{\textsc{Contrastive decoding} ($\CD$):} Ancestral sampling within $\vhead(\xstr)$ using logits $\CDsymb(\x\mid \xstr)$ (Eq.~\eqref{eq:CD_score}), which promote tokens preferred by $\pexp$ over $\pama$.
\end{enumerate}

We also study the effect of truncating the sampling support to further suppress low-probability continuations as follows:

\begin{enumerate}
    \setcounter{enumi}{2}
    \item \textbf{\NoContrast + $\vhead$:} Ancestral sampling from $\pexp(\cdot\mid \xstr)$ restricted to $\vhead(\xstr)$.
    \item \textbf{\NoContrast + top-$p$:} Ancestral sampling from $\pexp\left(\cdot\mid \xstr\right)$ restricted to top-$p$ selection \cite{holtzman_curious_2020}.
    \item \textbf{\NoContrast + top-$k$:} Ancestral sampling from $\pexp(\cdot\mid \xstr)$ restricted to top-$k$ selection \cite{fan_hierarchical_2018}.
    \item \textbf{$\CD$ with top-$p$:} Ancestral sampling restricted to the top-$p$ after already restricting to the $\vhead(\xstr)$ using logits $\CDsymb(\x\mid \xstr)$ (Eq.~\eqref{eq:CD_score}).
    \item \textbf{$\CD$ with top-$k$:} Ancestral sampling restricted to the top-$k$ after already restricting to the $\vhead(\xstr)$ using logits $\CDsymb(\x\mid \xstr)$ (Eq.~\eqref{eq:CD_score}).
\end{enumerate}

We sweep $k\in\{50,100,200\}$ and $p\in\{0.90,0.95,0.97\}$ and report effects on performance in \Cref{sec:truncation} and \Cref{tab:SummaryExperimentsShort}.

\subsection{The \bad and \good Models} \label{sec:amateur_model_families}
We consider three approaches to instantiate a \bad model $\pama$ (details in Appendix \ref{app:model_tok_details}): 

\begin{enumerate}[label=\roman*)]
    \item \textbf{Smaller models} that are $10\times$, $20\times$, $50\times$, and $100\times$ smaller than the \good model, and, following \cite{li_contrastive_2023}, selecting the checkpoint with the best evaluation perplexity. 
    \item \textbf{Earlier checkpoints}, e.g., if a \good checkpoint is taken at step 2500, we test \bad checkpoints at steps 2000, 1500, 1000 and 500.\looseness=-1
    \item \textbf{Attention dropout}, where the \bad model is the \good model, but run with attention dropout rates $\{0.1, 0.3, 0.5, 0.7\}$ at inference time \cite{phan_distillation_2024}.
\end{enumerate}

\paragraph{Note on scale.}
Prior evaluations of $\CD$ use billion-parameter \good models paired with much smaller \bad models (e.g., OPT-13B vs.\ OPT-125M; GPT-2-XL vs.\ GPT-2-small), and report that performance improves as the \good--\bad \emph{scale gap} increases \citep[§7.1; Fig.~2]{li_contrastive_2023}. 
$\CD$ is not limited to \good models that are several billion parameters or larger, e.g., as \citet{li_contrastive_2023} also show gains with GPT-2-XL (\(\sim\)1.5B).
However, the observed size-gap effect suggests that a strong contrast may be harder to elicit at our \(\sim\)100M-parameter scale. 
Consistent with this, \citet{obrien_contrastive_2023} find that smaller \bad models help more than larger ones and that gains tend to be stronger for larger \good models on reasoning tasks. 
We therefore investigate multiple \bad model instantiations to identify how we can elicit a sufficient contrastive signal at this scale \citep{li_contrastive_2023,obrien_contrastive_2023,phan_distillation_2024}.

\paragraph{Hyperparameters.}
Following \citet{li_contrastive_2023}, we use $\alpha = 0.1$ for $\vhead$ and the contrast strength is set to $\lambda = 1$.

\paragraph{The \good models.} We describe how the better models, $\pexp$, are selected in \Cref{sec:model_details_training}.

\subsection{Training with Mixed Corpora}

All models are trained from scratch to isolate the effect of the synthetic corpora. For each decoding method, we mix its 100M-token synthetic corpus with the same 100M-token TinyBabyLM corpus (see \Cref{sec:dataset}) used for the baselines, while keeping initialization seeds, training length, and optimization hyperparameters identical to the baseline runs. 
Batches contain 256 sequences of 1024 tokens with a fixed 70/30 mixture at the sequence level (\(70\%\) real, \(30\%\) synthetic) and are repeatedly regrouped and re-tokenized to act as a data regularizer (see Section \ref{sec:model_details_training} and Appendix ~\ref{app:model_tok_details}).

We ablate the original/synthetic mixture and report its effect on performance in Section~\ref{sec:_res_mixing_ratio}. 
Since initial testing indicated that the 70/30 mixture achieved the strongest average performance across tasks, we report results under this fixed ratio in the main experiments.

\section{Experimental Details}

\subsection{\emph{TinyBabyLM} Corpus}
\label{sec:dataset}

We start from the BabyLM 100M corpus and construct a modified variant by replacing the \textsc{Childes}, \textsc{BNC} and \textsc{Switchboard} portions with the synthetic \textit{TinyStories} \cite{eldan_tinystories_2023}. We add a portion of \textit{TinyStories} because \citet{eldan_tinystories_2023} show that their corpus, a constrained, child-directed synthetic corpus enables very small models (1–10M parameters) to learn fluent, grammatical multi-paragraph stories, making it a high-signal, data-efficient addition for low-resource pretraining. Concretely, we substitute \(\sim\)39.7M words of TinyStories for the removed words, yielding the following composition: Gutenberg (27.4M), SimpleWiki (14.9M), OpenSubtitles (17.7M) and TinyStories (39.7M) \cite{eldan_tinystories_2023,gerlach_standardized_2018,lison_opensubtitles2016_nodate}.
We refer to this modified corpus as \emph{TinyBabyLM}. The total amount of human-written+TinyStories text is held at \(\approx\)100M words; note that “words” here denote whitespace-delimited tokens, so totals differ from BPE token counts used during training (see Section~\ref{app:model_tok_details}).
We partition TinyBabyLM into three disjoint splits: \emph{train} (90.5M words), \emph{eval} (8.9M words), and \emph{seeds} (600K words). The generation seeds (a selection of $\sim$30K paragraph start prefixes) for synthetic generation are sampled exclusively from the \emph{seeds} split and are strictly disjoint from all \emph{train} and \emph{eval} text. To maintain balance across domains, the splits, \emph{seed}, \emph{train} and \emph{eval}, are distributed evenly across the four data sources.

\subsection{Model Architecture \& Training Setup}
\label{sec:model_details_training}
We use a decoder-only Transformer LLaMA-2 architecture with \(\sim\)100M parameters \cite{touvron_llama_2023}: 12 layers, hidden size 768, 12 attention heads, MLP intermediate size 3072, and a maximum context length of 1024 tokens. All models are trained from scratch with the same initialization scheme.\looseness-1

Tokenization is performed with a SentencePiece BPE tokenizer (vocabulary size 32k) trained on the TinyBabyLM corpus; the same tokenizer is used for all experiments to ensure comparability (see Appendix~\ref{app:model_tok_details} for details).

Training uses a global batch of 256 sequences \(\times\) 1024 tokens, AdamW with weight decay 0.1, and a cosine learning-rate schedule: peak \(1\times10^{-3}\), 150 warm-up steps, and decay to zero by step 8000. The training duration is fixed to 8000 steps for every run and checkpoints are saved every 500 steps. Each experimental condition is repeated with \(n=10\) distinct random seeds.

\paragraph{Data pipeline (applies to all runs).}
Real and synthetic corpora are stored as rows of text and, at the start of training, are independently shuffled, tokenized, and split into fixed-length sequences. Sampling proceeds until a corpus is exhausted, at which point that corpus is reshuffled, and re-segmented before resuming. This periodic resegmentation acts as a regularizer and is applied identically to baseline and mixed-data runs.

\paragraph{Good checkpoint selection.}
From each of the \(n=10\) \Baseline seeds, we first select the saved checkpoint with the lowest perplexity, forming the candidate set \(\mathcal{X}\). We then evaluate only \(\mathcal{X}\) on the full suite of tasks, convert scores to percentiles within the task, average percentile across tasks, and choose as the \good model the checkpoint with the highest average percentile.

\subsection{Evaluation \& Statistical Analysis}
\label{sec:eval_statistics}
\paragraph{Benchmarks.}
We evaluate on the zero-shot BabyLM evaluation suite\footnote{\href{https://github.com/babylm/evaluation-pipeline-2025}{https://github.com/babylm/evaluation-pipeline-2025}} and report Perplexity on the TinyBabyLM \emph{eval}-split (see ~\ref{sec:dataset}). The tasks considered are:

\begin{itemize}[leftmargin=1em,labelsep=0.5em]
  \item \textbf{BLiMP}: Benchmark of Linguistic Minimal Pairs testing core English grammar linguistic competence \cite{warstadt_blimp_2023}.
  \item \textbf{BLiMP Supplement}: BLiMP-style suite, extending to dialogue and question answering, focused on reasoning, syntax and semantics \cite{hu_findings_2024,warstadt_findings_2023}.
  \item \textbf{EWoK}: Checks for social/physical/world knowledge and semantic understanding \cite{ivanova_elements_2024}.\looseness=-1
  \item \textbf{Entity Tracking}: Requires maintaining and updating entity states across text to test memory and state reasoning \cite{kim_entity_2023}.
  \item \textbf{WUG}: Evaluates morphology, evaluating on adjective nominalization to estimate linguistic generalization \cite{hofmann_derivational_2024}.
  \item \textbf{Reading}: Compares model surprisal to human word-by-word reading times to assess processing alignment \cite{de_varda_cloze_2023}.
  \item \textbf{Eye-Tracking}: Tests whether model predictability tracks human eye-movement measures during reading \cite{de_varda_cloze_2023}.
\end{itemize}

The metric used for the \textbf{Reading} and \textbf{Eye-tracking} tasks is the partial change (\%) in the coefficient of determination, that is, the additional proportion of variance explained. For the other tasks, accuracy is used.

\tableBaselineExpert
\paragraph{Per-task mean-max over checkpoints.}
For each training method\footnote{Either \Baseline, or a pair of decoding strategy  from \ref{sec:decoding_strategies} and bad model setting from \ref{sec:amateur_model_families}} $m$, benchmark task $t$, and initialization seed $s$, we save checkpoints every 500 steps and select the best checkpoint independently per $(m,t,s)$. Let $\mathcal{C}_{m,t,s}$ denote the set of saved checkpoints over the steps, and $S_{m,t,s}(c)$ the task score at checkpoint $c$. For higher-is-better tasks, we set
\[
c^*_{m,t,s}\defeq \arg\max_{c\in\mathcal{C}_{m,t,s}} S_{m,t,s}(c),
\]
while for perplexity we take $\arg\min$. The selected checkpoint $c^*_{m,t,s}$ is then evaluated. This procedure estimates the best attainable performance per task under the fixed training budget and avoids coupling to a single global checkpoint.

\paragraph{Paired bootstrap for statistical significance.}
Evaluation of checkpoint $c^*_{m,t,s}$ for task t yields per-example outcomes $y_{m,t,s,i}$ for examples $i=1,\dots,N_{t}$. We use paired bootstrap with $B=1000$ resamples to calculate confidence intervals. For each $(t,s)$ and bootstrap draw $b$, sample the index-set $I^{(b)}$ of size $N_{t}$ with replacement from $\{1,\dots,N_{t}\}$ and apply the \emph{same} $I^{(b)}$ to all methods (pairing). We average out uncertainty over the seeds:\looseness=-1
\begin{align}
   \bar{y}^{(b)}_{m,t,s} &=\frac{1}{N_t}\sum_{i\in I^{(b)}} y_{m,t,s,i} \label{eq:mean_tasks}\\
    \mu^{(b)}_{m,t}&= \frac{1}{|\mathcal{S}|}\sum_{s\in\mathcal{S}} \bar{y}^{(b)}_{m,t,s}. 
\end{align}

As in \eqref{eq:mean_tasks}, $\bar{y}^{(b)}_{m,t,s}$ is the mean for tasks with per-example scalar scores (e.g., BLiMP, EWoK). For metrics with task-specific aggregations (e.g., Perplexity or Reading), we substitute the appropriate aggregation function and proceed identically. 
For a comparison of two methods $m_1$ and $m_2$, we form the bootstrap difference distribution
\begin{equation}   
\Delta^{(b)}_t=\mu^{(b)}_{m_1,t}-\mu^{(b)}_{m_2,t}
\end{equation}
We compute 95\% confidence intervals via the percentile method, $\mathrm{CI}_{95} = [\mathrm{pct}_{2.5},\,\mathrm{pct}_{97.5}]$ of $\{\Delta^{(b)}_t\}_{b=1}^{B}$. A difference is deemed significant if $0\notin \mathrm{CI}_{95}$.
We compute one-sided $p$-values in the direction of the observed effect using the estimator on the bootstrap differences $\{\Delta_t^{(b)}\}_{b=1}^B$: for higher-is-better tasks with $\hat{\Delta}_t>0$,
\begin{align}
    p=\frac{1+\sum_{b=1}^{B}\mathbb{I}\{\Delta_t^{(b)}\le 0\}}{B+1} \label{eq:p_value_defn}
\end{align}
and if $\bar{\Delta}_t<0$ use $\ge$ instead. For lower-is-better metrics we swap the inequality accordingly.

\paragraph{Aggregate reporting.}
For tables and figures, we bold the best method per benchmark and mark significant improvements/degradations relative to the \Baseline. We report, for each method $m$ and task $t$, the bootstrap mean $\bar{\mu}_{m,t}$ and standard-error.
\begin{align}
    \bar{\mu}_{m,t} \;=\; \frac{1}{B}\sum_{b=1}^{B}\mu^{(b)}_{m,t}, \quad \widehat{\mathrm{SE}}_{m,t}= \frac{\sigma_{m,t}}{\sqrt{B}} \label{eq:defn_mean_std}
\end{align}
This analysis serves to estimate the maximum achievable performance for each method, on each task, given the training setup.
Our aggregating metric \mudeltarel is the mean relative performance, across all tasks except Perplexity, vs. the \Baseline---i.e., it is the average proportional change given in percentages.\looseness=-1

\section{Results}

\subsection{\Baseline Performance}

\Cref{tab:BaselineExpert} summarizes the performance of our reference points, the \good and \Baseline results. Recall that the \Baseline row reports the mean $\pm$ s.e.\ over $n{=}10$ independent runs under our per-task bootstrapped mean–max evaluation (\Cref{sec:eval_statistics}). In contrast, the \good model is a \emph{single} checkpoint selected once, across seeds, using the selection procedure described in \Cref{sec:model_details_training}. As such $\good$ is broadly representative of a strong model but sits slightly below the \Baseline mean on some tasks (e.g., Perplexity, BLiMP Supplement) because it cannot adapt per task. We use this fixed checkpoint as the \good model in all subsequent synthetic corpora and comparisons.

\subsection{Contrastive vs. non-contrastive generation}
\label{sec:results_contrastive_overview}

\tableSummaryExperimentsSmallNoRel

\paragraph{Setup.}
Recall from \Cref{sec:decoding_strategies} that we compare the three generation settings for synthesizing the 100M-token corpora: (i) \NoContrast, (ii) \NoContrastVHead, and (iii) $\CD$ . Among all contrastive instantiations, using the early checkpoint at 500 steps (\textbf{CD–Early–500}) emerged as the strongest (see \Cref{sec:search-cd-settings}), and we use it as our $\CD$ representative in this section. Results are summarized in \Cref{tab:SummaryExperimentsSmallNoRel}.

\paragraph{Aggregate performance.}
All synthetic regimes beat \Baseline. $\CD$ delivers the strongest overall gains (\mudeltarel +4.90\%), with the non-contrastive variants lacking, see \Cref{tab:SummaryExperimentsSmallNoRel}.

\paragraph{Language modeling (Perplexity).}
Perplexity drops for every method. \NoContrast attains the lowest value (23.56), with $\CD$ close behind, so non-contrastive sampling edges out $\CD$ slightly on the LM objective, while $\CD$ still clearly improves over \Baseline; see \Cref{tab:SummaryExperimentsSmallNoRel}.

\tableCDEarlyVsNoContrastDeltaTransposedSign

\paragraph{Task-level pattern and head-to-head.}
$\CD$ performs best on five tasks and shows significant gains on five, notably on reasoning-/tracking-oriented evaluations like BLiMP Supplement, Entity Tracking, and EWoK (see \Cref{tab:SummaryExperimentsSmallNoRel}. In contrast, \NoContrast is best on three tasks with significant effects on four, and it leads on core linguistic competence with Perplexity and BLiMP. In direct statistical comparisons (\CD\ vs. \NoContrast), as displayed in \Cref{tab:CDEarlyVsNoContrastDeltaTransposed}, \NoContrast has a small but significant edge on Perplexity and BLiMP, whereas $\CD$ achieves significant, and generally larger, gains on Entity Tracking, EWoK, and WUG. The remaining tasks show no reliable difference.

\paragraph{Is it the $\vhead$ mask or the contrastive logits?}
\NoContrastVHead serves as a control that isolates the effect of restricting to the $\alpha$-head without any contrastive subtraction. If head-masking alone explained $\CD$’s gains, \NoContrastVHead would mirror $\CD$. It does not: while \NoContrastVHead modestly helps Perplexity (–0.51\%) and Eye Tracking (+12.12\%), it significantly hurts Entity Tracking (–8.45\%) and yields small/neutral changes elsewhere (\Cref{tab:SummaryExperimentsSmallNoRel}). This suggest that the improvements are driven by the contrastive logits and not the $\vhead$ constraint.

\paragraph{Takeaway.}
Mixing synthetic data consistently helps. Among generation strategies, $\CD$  delivers the strongest overall improvements and a clear advantage on reasoning-oriented benchmarks, while \NoContrast remains best for the LM objective and BLiMP. The \NoContrastVHead control suggests that contrastive scoring, not head-masking, is the key to $\CD$’s benefits.

\tableSummaryExperimentsShort   

\subsection{Searching for Effective \CD\ Settings}
\label{sec:search-cd-settings}

We instantiate the amateur for contrastive decoding using three settings (\Cref{sec:amateur_model_families}): (i) earlier checkpoints, (ii) smaller models, and (iii) inference-time attention dropout. We report the best setting from each setting in \Cref{tab:SummaryExperimentsShort} and the full results in \Cref{tab:AllModels} in the Appendix. 
While all $\CD$ versions give some boost in performance, using an earlier checkpoint gives the strongest signal.

\subsection{Effect of truncation}
\label{sec:truncation}

\begin{figure}
    \centering
    \includegraphics[width=1.0\linewidth]{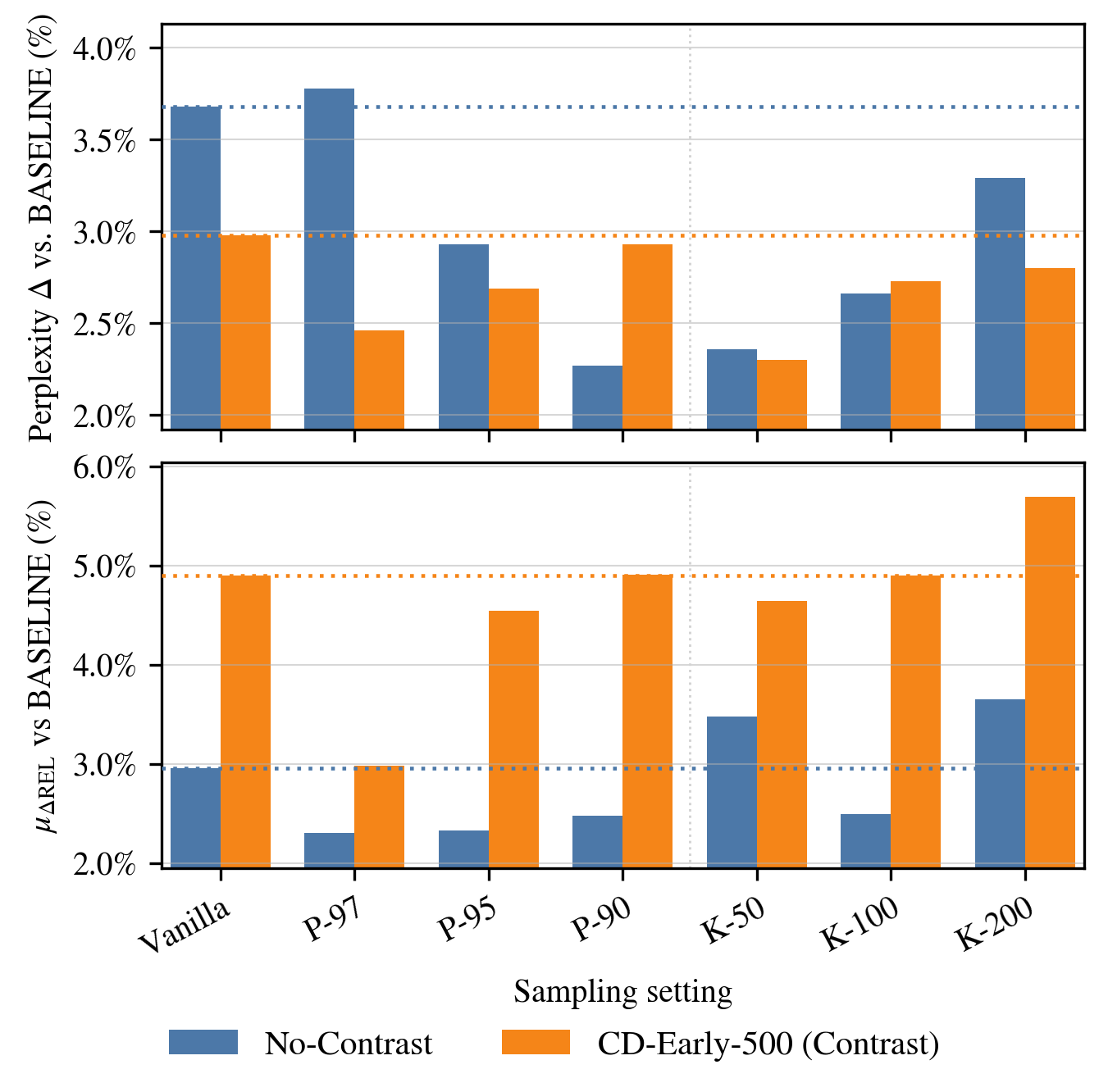}
    \caption{Top-$k$ and top-$p$ truncation under ancestral decoding. ``Vanilla'' denotes ancestral sampling from unmodified logits after \CD{} or \NoContrast. On downstream tasks, $k{=}200$ is the strongest setting; perplexity exhibits no single optimum. Full results in \Cref{tab:AllModels}.}

    \label{fig:toppkplot}
\end{figure}

Across both non-contrastive and contrastive generators, truncation yields at most modest gains, see \Cref{fig:toppkplot} and \Cref{tab:SummaryExperimentsShort}, for the full sweep \Cref{tab:AllModels}. Benefits are largest for Top-$k$ with $k=200$; nucleus truncation is less reliable.

\textsc{CD--Early--500--Top-k--200} attains the best aggregate improvement, increasing $\mudeltarel$ to $5.69\%$ (vs.\ $4.90\%$ for \textsc{CD--Early--500}) at essentially unchanged perplexity (23.77 vs.\ 23.73). Slightly tighter truncation with \textsc{CD--Early--500--Top-k--100} delivers the strongest \emph{Entity Tracking} ($+19.02\%$) and the best \emph{EWoK} ($+1.44\%$), indicating that modest tail pruning can amplify the contrastive signal, with small trade-offs on \emph{Reading Alignment} and \emph{WUG}.

For \NoContrast, nucleus truncation marginally improves perplexity but reduces $\mudeltarel$. In contrast, \textsc{\NoContrast-Top-k–200} raises $\mudeltarel$ to $3.65\%$ while reducing perplexity.

Within our (limited) sweep, truncation can provide additional headroom, especially for contrastive decoding. Light Top-$k$ (\,$k\!\in\![100,200]$\,) appears to preserve diversity while reinforcing preferences for higher-signal tokens.

\subsection{Mixing Ratio Ablation}
\label{sec:_res_mixing_ratio}
\begin{figure}
    \centering
    \includegraphics[width=1\linewidth]{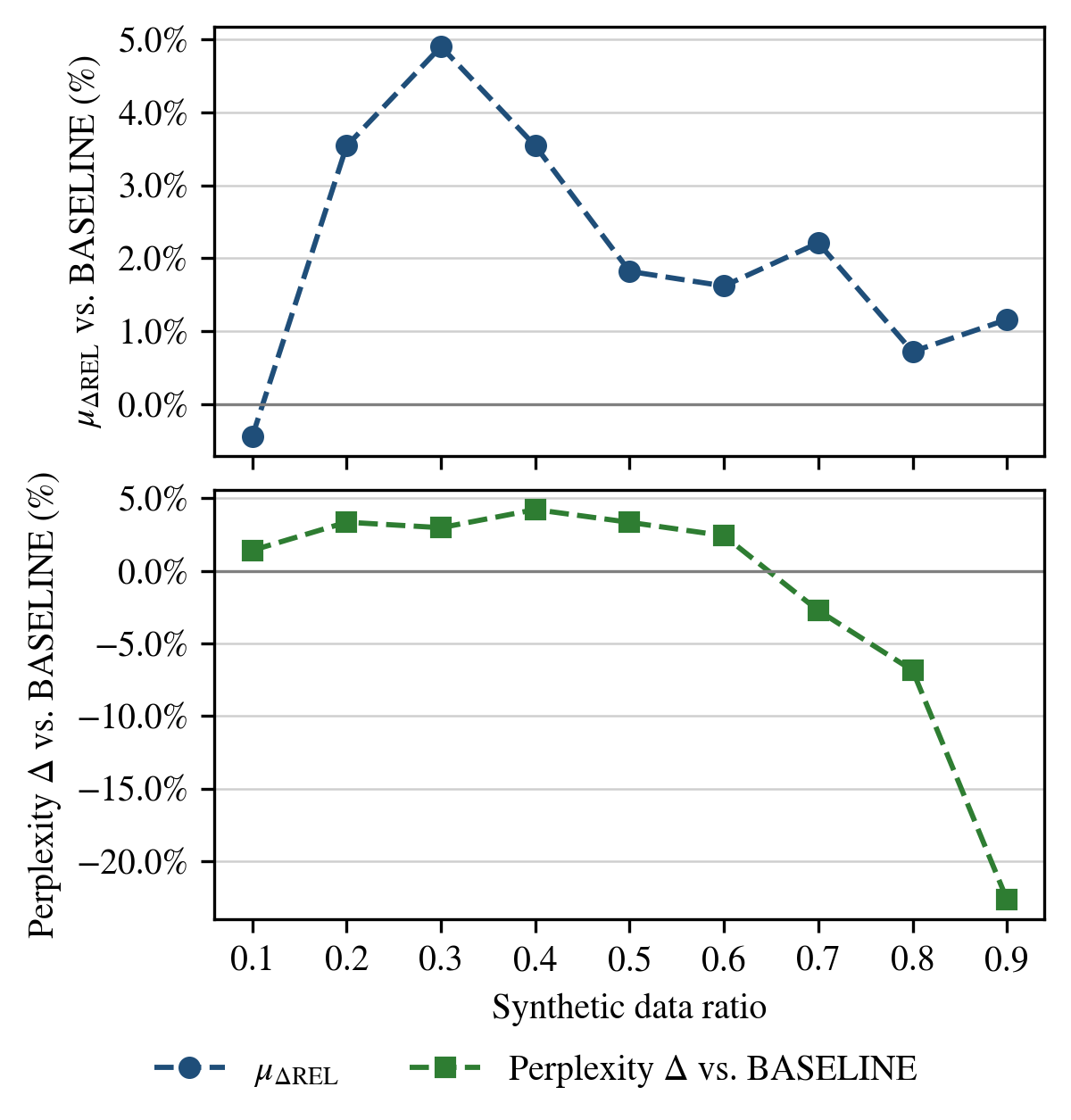}
    \caption{Mixing ratio ablation for CD-generated synthetic corpora (CD-Early-500), also see in \Cref{tab:AllModels}. The ratio indicates the fraction of synthetic data in training batches. \mudeltarel{} is the mean relative improvement over \Baseline{} across non-perplexity tasks; Perplexity shows relative change vs.\ \Baseline{};. A 30\% mix yields the best overall \mudeltarel{} (+4.90\%), while 40\% attains the lowest perplexity (23.42).}
    \label{fig:mixing_ratio_plot}
\end{figure}

We analyze what proportion of the original and $\CD$-generated data is most beneficial by varying their ratio. The results can be seen in \Cref{fig:mixing_ratio_plot}. Note that all corpora were generated with the CD-Early-500 setting. The full result are shown in the Appendix in \Cref{tab:AllModels}. A ratio of 30\% synthetic data performs best. Interestingly, similar ratios have shown to perform well when including semi-synthetic data in machine translation using back-translations \cite{fadaee_back-translation_2018, simonarson_mieinds_nodate}.

\section{Discussion}

This work asks whether inference-time \CD\ can be repurposed as a \emph{corpus generator} for improving pre-training of language-models. Three findings stand out.

\paragraph{Mixing synthetic data helps; \CD\ helps most where reasoning is required.}
Across the BabyLM suite, adding any synthetic corpus to TinyBabyLM improves over the \Baseline trained only on real text (\Cref{tab:SummaryExperimentsSmallNoRel}). Among generators, \CD\ delivers the strongest aggregate gains (\mudeltarel $+4.90\%$ for standard sampling and $+5.69\%$ using top-$k$) and the clearest advantages on reasoning- and tracking-oriented tasks (BLiMP Supplement, Entity Tracking, EWoK, WUG). By contrast, non-contrastive sampling yields the lowest Perplexity and leads on BLiMP, suggesting it better reinforces core grammatical regularities. Together, these results support a practical division of labor: use \CD\ when downstream targets emphasize multi-step inference, state maintenance, or world knowledge; use vanilla sampling when the objective is to minimize perplexity or to improve core grammaticality. A combined approach could also be considered.

\paragraph{Contrastive \emph{scoring}, not head masking, is the key ingredient.}
The \NoContrastVHead control, which applies only the $\alpha$-head mask from the good model, does not replicate \CD's benefits and can even hurt Entity Tracking. This indicates that the subtraction against a worse model is doing the heavy lifting. Intuitively, \CD\ preserves high-plausibility tokens while attenuating those over-predicted by the amateur, reducing topical drift and shallow heuristics that smaller or earlier checkpoints tend to prefer effects that plausibly matter most for reasoning-heavy benchmarks.

\paragraph{A practical amateur: earlier checkpoints are a strong and simple choice.}
Among amateur families, an earlier checkpoint of the same architecture (\textsc{CD–Early–500}) performs best in our sweep (\Cref{tab:AllModels}). This choice is attractive operationally: it requires no additional model training, and produced a non-trivial contrast. Smaller-model amateurs and dropout-only amateurs also work but did not perform as well.

\paragraph{Broader implications.}
These results suggest that inference-time guidance can be re-purposed into \emph{corpus-level} signal shaping: by subtracting the preferences of a systematically weaker model, the generator appears to skew synthetic text toward trajectories that contain constraints that more relevant for reasoning tasks.

\section{Limitations}

\paragraph{Scale and budget.}
All experiments use $\sim$100M-parameter models, a fixed 8k-step budget, and an English-only, curated TinyBabyLM corpus. Findings may not transfer to larger scales, non-English data, or web-scale pre-training.

\paragraph{Amateur choice and hyperparameters.}
Although multiple amateur families were explored, the sweep is not exhaustive. The strongest setting (\textsc{Early–500}) may depend on save frequency, optimizer dynamics, or data order. We kept $\alpha{=}0.1$ and $\lambda{=}1$ fixed.

\paragraph{Compute and memory overhead.}
\CD\ generation requires concurrent access to both expert and amateur models at inference time, roughly doubling activation memory and increasing generation latency. While dropout-based amateurs reduce memory pressure, they did not consistently match the early-checkpoint amateur in our setting.

\paragraph{Distributional narrowing.}
Head masking constrains support and can reduce lexical diversity; while \CD\ outperformed the head-only control, the mask remains part of the procedure, which may under-represent rare constructions. Effects on long-tail generalization and stylistic diversity were not directly measured.

\paragraph{Safety, bias, and factuality.}
No human evaluation of safety or factual correctness was conducted, and no targeted bias audits were performed. Although \CD\ can downweight some amateur-preferred artifacts, it may also amplify biases present in the expert. More rigorous filtering and auditing are needed for deployment-facing settings.

\paragraph{Single iteration.} We only consider a single iteration of $\CD$, in follow-up work we plan to consider how repeated application of $\CD$ scales.

\section*{Acknowledgments}
Vésteinn Snæbjarnarson is supported by the Pioneer Centre for AI, DNRF grant number P1.

\bibliography{Citations_MS_Thesis_Proposal}

\appendix
\section*{Appendix}
\label{sec:appendix}

\section{Model \& Tokenizer Training Details}
\label{app:model_tok_details}

\paragraph{Model Details}
The architecture we use is a LLaMA-2–style decoder-only Transformer from \citet{touvron_llama_2023} with \texttt{name=llama-12-768}: 12 layers, hidden size 768, 12 attention heads, MLP intermediate size 3072, and maximum context length 2048 tokens. All models use \texttt{dtype=float32} and the same tokenizer configuration.

\paragraph{Tokenizer.}
We use a SentencePiece BPE tokenizer (vocabulary size 32{,}000) trained on the TinyBabyLM corpus. Preprocessing follows SentencePiece defaults, including Unicode normalization, whitespace deduplication, and removal of control characters. The identical tokenizer is used across all experiments to ensure comparability.

\paragraph{Training Details, Data Mixing \& Regrouping Regularizer.}
Per-device batches contain 16 sequences of length 1{,}024 tokens; with 4 GPUs and gradient accumulation of 4, the effective global batch is \(256 \times 1{,}024\) tokens.

For training with a (\(70\%\) real, \(30\%\) synthetic) mixture for each batch, the 256 sequences are sampled from the original/synthetic corpus accordingly to satisfy the required ratio at a sequence-ratio level. To implement this mixture, the real and synthetic corpora are stored as rows of text and, at the start of training, each corpus is independently shuffled, tokenized, and split into fixed-length sequences. Sequences are then sampled until one corpus is exhausted; the exhausted corpus is reshuffled, re-tokenized, and re-split before sampling resumes. This periodic resegmentation acts as a light regularizer by continually refreshing ordering and boundaries, and we apply the identical procedure in the \Baseline runs for parity

We train with the causal language modeling objective (next-token prediction), minimizing token-level cross-entropy (negative log-likelihood). Optimization uses AdamW with \(\beta_1=0.9\), \(\beta_2=0.999\), weight decay 0.1, and initial learning rate \(1\text{e}{-3}\). The schedule is cosine with 150 warm-up steps, decaying to zero by step 8{,}000. All runs are executed on a multi-GPU cluster with NVIDIA RTX 3090 or RTX 4090 GPUs.

\begin{table}[t]
\centering
\caption{Architectures used for the good and bad models. All models share the same tokenizer and max position embeddings (1024). The suffix in \texttt{name} (e.g., \texttt{5x}, \texttt{10x}) indicates the intended scale relative to the expert.}
\label{tab:amateur-configs}
\adjustbox{width=\linewidth}{%
\begin{tabular}{lrrrrr}
\toprule
\textbf{Name} & \textbf{Layers} & \textbf{Hidden} & \textbf{Heads} & \textbf{Intermediate} & \textbf{Max pos} \\
\midrule
\texttt{llama-12-768} ($\good$)     & 12 & 768 & 12 & 3072 & 1024 \\
\texttt{llama-10-512-5x}           & 10 & 512 &  8 & 2048 & 1024 \\
\texttt{llama-8-384-10x}           &  8 & 384 &  6 & 1536 & 1024 \\
\texttt{llama-6-256-20x}           &  6 & 256 &  4 & 1024 & 1024 \\
\texttt{llama-5-224-50x}           &  5 & 224 &  4 &  896 & 1024 \\
\texttt{llama-4-192-100x}          &  4 & 192 &  3 &  768 & 1024 \\
\bottomrule
\end{tabular}}

\end{table}

\section{Synthetic Generation Details}
\label{app:synth_gen_details}

\paragraph{Framework and hardware.}
Synthetic text is produced with a custom, PyTorch generation loop designed for efficiency and flexibility. The loop supports multi-GPU parallelization, per-token logit transforms (for contrastive decoding), and caching. All generation runs on the same multi-GPU cluster used for training, typically \(4\times\) NVIDIA RTX~3090/4090 GPUs.

\section{All Task Results}
We give a comprehensive overview of model performance in \Cref{tab:AllModels}.
\tableAllModels

\end{document}